\documentclass[10pt,twocolumn,letterpaper]{article}

\usepackage{titling}
\usepackage{cvpr}
\usepackage{times}
\usepackage{enumitem}
\usepackage{subfig}
\usepackage{amsmath}
\usepackage{amssymb}
\usepackage{mathtools}
\usepackage{makecell}
\usepackage{multirow}
\usepackage{balance}

\usepackage[breaklinks=true,bookmarks=false]{hyperref}

\cvprfinalcopy 

\def\cvprPaperID{116} 

\ifcvprfinal\fi
\begin{document}

\title{Human-centric Indoor Scene Synthesis Using Stochastic Grammar}

\author{
\vspace{-6pt}
Siyuan Qi$^{1}$ \quad Yixin Zhu$^{1}$ \quad Siyuan Huang$^{1}$ \quad Chenfanfu Jiang$^{2}$ \quad  Song-Chun Zhu$^1$\\[12pt]
$^1$ UCLA Center for Vision, Cognition, Learning and Autonomy\\
$^2$ UPenn Computer Graphics Group
\vspace{-6pt}
}

\maketitle

\begin{abstract}
We present a human-centric method to sample and synthesize 3D room layouts and 2D images thereof, to obtain large-scale 2D/3D image data with perfect per-pixel ground truth. 
An attributed spatial And-Or graph (S-AOG) is proposed to represent indoor scenes. The S-AOG is a probabilistic grammar model, in which the terminal nodes are object entities. Human contexts as contextual relations are encoded by Markov Random Fields (MRF) on the terminal nodes.
We learn the distributions from an indoor scene dataset and sample new layouts using Monte Carlo Markov Chain.
Experiments demonstrate that our method can robustly sample a large variety of realistic room layouts based on three criteria: (i) visual realism comparing to a state-of-the-art room arrangement method, (ii) accuracy of the affordance maps with respect to ground-truth, and (ii) the functionality and naturalness of synthesized rooms evaluated by human subjects. The code is available at \url{https://github.com/SiyuanQi/human-centric-scene-synthesis}.
\end{abstract}

\section{Introduction}\label{sec:intro}

Traditional methods of 2D/3D image data collection and ground-truth labeling have evident limitations. i) High-quality ground truths are hard to obtain, as depth and surface normal obtained from sensors are always noisy. ii) It is impossible to label certain ground truth information, \eg, 3D objects sizes in 2D images. iii) Manual labeling of massive ground-truth is tedious and error-prone even if possible. To provide training data for modern machine learning algorithms, an approach to generate large-scale, high-quality data with the perfect per-pixel ground truth is in need.


\begin{figure}[t!]
	\begin{center}
		\includegraphics[width=\linewidth]{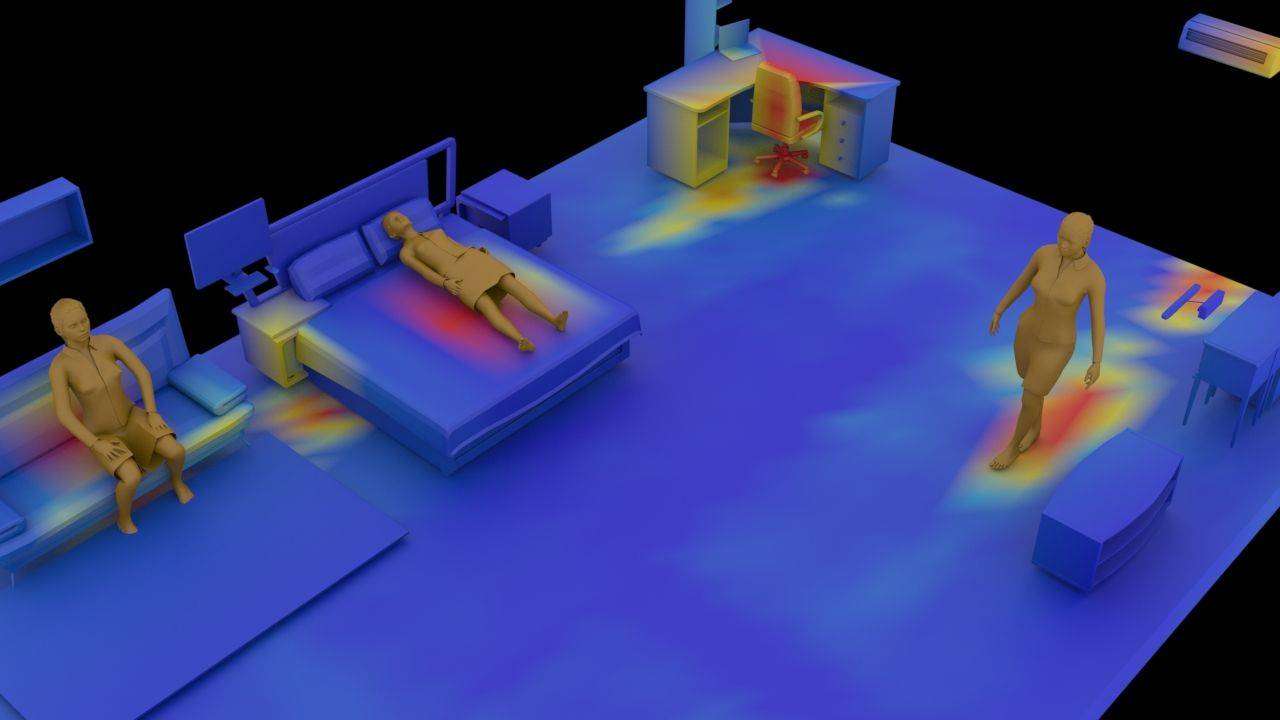}
	\end{center}
	\vspace{-10pt}
	\caption{An example of synthesized indoor scene (bedroom) with affordance heatmap. The joint sampling of a scene is achieved by alternative sampling of humans and objects according to the joint probability distribution.}
	\vspace{-12pt}
	\label{fig:overview}
\end{figure}

In this paper, we propose an algorithm to automatically generate a large-scale 3D indoor scene dataset, from which we can render 2D images with pixel-wise ground-truth of the surface normal, depth, and segmentation, \etc. The proposed algorithm is useful for tasks including but not limited to: i) learning and inference for various computer vision tasks; ii) 3D content generation for 3D modeling and games; iii) 3D reconstruction and robot mappings problems; iv) benchmarking of both low-level and high-level task-planning problems in robotics.

\begin{figure*}[t!]
	\begin{center}
		\includegraphics[width=\linewidth]{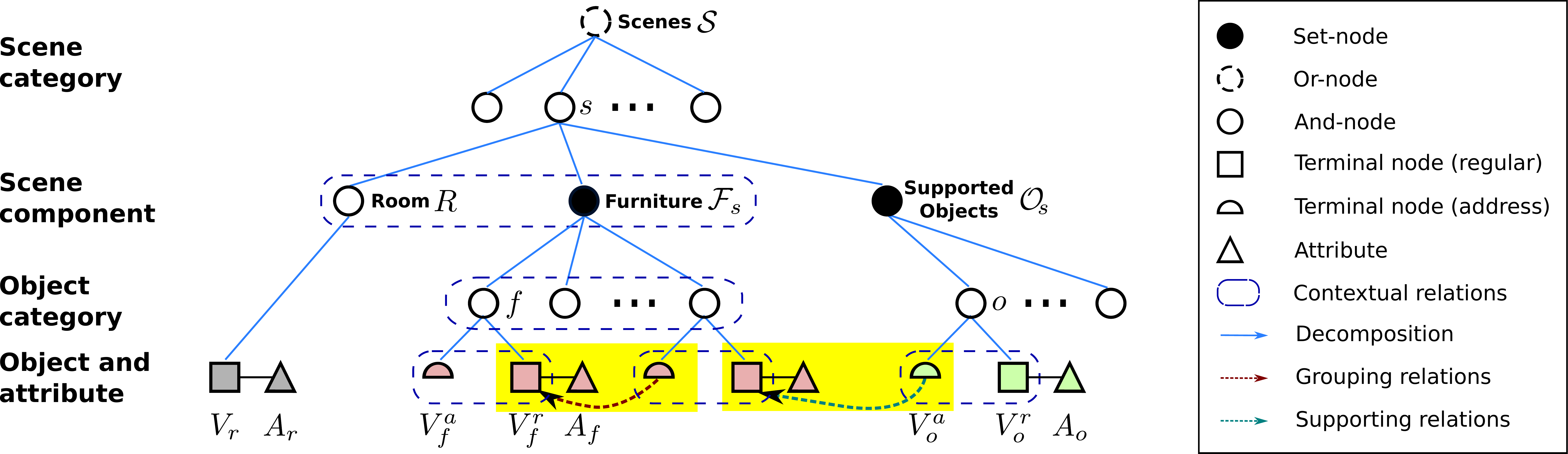}
	\end{center}
	\vspace{-10pt}
	\caption{Scene grammar as an attributed S-AOG. A scene of different types is decomposed into a room, furniture, and supported objects. Attributes of terminal nodes are internal attributes (sizes), external attributes (positions and orientations), and a human position that interacts with this entity. Furniture and object nodes are combined by an address terminal node and a regular terminal node. A furniture node (\eg, a chair) is grouped with another furniture node (\eg, a desk) pointed by its address terminal node. An object (\eg, a monitor) is supported by the furniture (\eg, a desk) it is pointing to. If the value of the address node is null, the furniture is not grouped with any furniture, or the object is put on the floor. Contextual relations are defined between the room and furniture, between a supported object and supporting furniture, among different pieces of furniture, and among functional groups.}
	\label{fig:representation}
	\vspace{-12pt}
\end{figure*}


Synthesizing indoor scenes is a non-trivial task. It is often difficult to properly model either the relations between furniture of a functional group, or the relations between the supported objects and the supporting furniture. Specifically, we argue there are four major difficulties. (i) In a functional group such as a dining set, the number of pieces may vary. 
(ii) Even if we only consider pair-wise relations, there is already a quadratic number of object-object relations. (iii) What makes it worse is that most object-object relations are not obviously meaningful. For example, it is unnecessary to model the relation between a pen and a monitor, even though they are both placed on a desk. (iv) Due to the previous difficulties, an excessive number of constraints are generated. Many of the constraints contain loops, making the final layout hard to sample and optimize.

To address these challenges, we propose a human-centric approach to model indoor scene layout. It integrates human activities and functional grouping/supporting relations as illustrated in Figure~\ref{fig:overview}. This method not only captures the human context but also simplifies the scene structure. Specifically, we use a probabilistic grammar model for images and scenes~\cite{zhu2007stochastic} -- an attributed spatial And-Or graph (S-AOG), including vertical hierarchy and horizontal contextual relations. The contextual relations encode functional grouping relations and supporting relations modeled by object affordances~\cite{gibson1979ecological}. For each object, we learn the affordance distribution, \ie, an object-human relation, so that a human can be sampled based on that object. Besides static object affordance, we also consider dynamic human activities in a scene, constraining the layout by planning trajectories from one piece of furniture to another. 

In Section~\ref{sec:representation}, we define the grammar and its parse graph which represents an indoor scene. We formulate the probability of a parse graph in Section~\ref{sec:formulation}. The learning algorithm is described in Section~\ref{sec:learning}. Finally, sampling an indoor scene is achieved by sampling a parse tree (Section~\ref{sec:synthesis}) from the S-AOG according to the prior probability distribution. 

This paper makes three major \textbf{contributions}. (i) We jointly model objects, affordances, and activity planning for indoor scene configurations. (ii) We provide a general learning and sampling framework for indoor scene modeling. (iii) We demonstrate the effectiveness of this structured joint sampling by extensive comparative experiments.

\subsection{Related Work}\label{sec:related_work}

\textbf{3D content generation} is one of the largest communities in the game industry and we refer readers to a recent survey~\cite{hendrikx2013procedural} and book~\cite{shaker2016procedural}. In this paper, we focus on approaches related to our work using probabilistic inference. Yu~\cite{yu2011make} and Handa~\cite{handascenenet} optimize the layout of rooms given a set of furniture using MCMC, while Talton~\cite{talton2011metropolis} and Yeh~\cite{yeh2012synthesizing} consider an open world layout using RJMCMC. These 3D room re-arrangement algorithms optimize room layouts based on constraints to generate new room layouts using a given set of objects. In contrast, the proposed method is capable of adding or deleting objects without fixing the number of objects.
Some literature focused on fine-grained room arrangement for specific problems, \eg, small objects arrangement using user-input examples~\cite{fisher2012example} and procedural modeling of objects to encourage volumetric similarity to a target shape~\cite{ritchie2015controlling}. 
To achieve better realism, Merrell~\cite{merrell2011interactive} introduced an interactive system providing suggestions following interior design guidelines. Jiang~\cite{jiang2016modeling} uses a mixture of conditional random field (CRF) to model the hidden human context and arrange new small objects based on existing furniture in a room. However, it cannot direct sampling/synthesizing an indoor scene, since the CRF is intrinsically a discriminative model for structured classification instead of generation.

\textbf{Synthetic data} has been attracting an increasing interest to augment or even serve as training data for object detection and correspondence~\cite{dwibedi2017cut,mccormac2017scenenet,qi2016volumetric,song2014sliding,sun2014virtual,zhang2016physically,zhou2016learning}, single-view reconstruction~\cite{huang2015single}, pose estimation~\cite{chen2016synthesizing,shakhnarovich2003fast,su2015render,yasin2016dual}, depth prediction~\cite{su2014estimating}, semantic segmentation~\cite{richter2016playing}, scene understanding~\cite{handa2016understanding,handascenenet,zhang2016deepcontext}, autonomous pedestrians and crowd~\cite{ondvrej2010synthetic,qi2018intent,shao2005autonomous}, VQA~\cite{johnson2016clevr}, training autonomous vehicles~\cite{chen2015deepdriving,dosovitskiy17carla,MSR-TR-2017-9}, human utility learning~\cite{ye2017martian,zhu2016inferring} and benchmarks~\cite{handa2014benchmark,qiu2016unrealcv}.


\textbf{Stochastic grammar model} has been used for parsing the hierarchical structures from images of indoor~\cite{liu2014single,zhao2013scene} and outdoor scenes~\cite{liu2014single}, and images/videos involving humans~\cite{qi2017predicting,wang2018attentive}. In this paper, instead of using stochastic grammar for parsing, we forward sample from a grammar model to generate large variations of indoor scenes.

\section{Representation of Indoor Scenes}\label{sec:representation}
We use an attributed S-AOG~\cite{zhu2007stochastic} to represent an indoor scene. An attributed S-AOG is a probabilistic grammar model with attributes on the terminal nodes. It combines i) a probabilistic context free grammar (PCFG), and ii) contextual relations defined on an Markov Random Field (MRF), \ie, the horizontal links among the nodes. The PCFG represents the hierarchical decomposition from scenes (top level) to objects (bottom level) by a set of terminal and non-terminal nodes, whereas contextual relations encode the spatial and functional relations through horizontal links. The structure of S-AOG is shown in Figure~\ref{fig:representation}.


Formally, an S-AOG is defined as a 5-tuple: $\mathcal{G} = \langle S, V, R, P, E \rangle$, where we use notations $S$ the root node of the scene grammar, $V$ the vertex set, $R$ the production rules, $P$ the probability model defined on the attributed S-AOG, and $E$ the contextual relations represented as horizontal links between nodes in the same layer. \footnote{We use the term ``vertices" instead of ``symbols" (in the traditional definition of PCFG) to be consistent with the notations in graphical models.}

\textbf{Vertex Set} $V$ can be decomposed into a finite set of non-terminal and terminal nodes: $V = V_{NT} \cup V_{T}$.
\begin{itemize}[wide,leftmargin=0cm,noitemsep,nolistsep]
	\item $V_{NT} = V^{And} \cup V^{Or} \cup V^{Set}$. The non-terminal nodes consists of three subsets. i) A set of \textbf{And-nodes} $V^{And}$, in which each node represents a decomposition of a larger entity (\eg, a bedroom) into smaller components (\eg, walls, furniture and supported objects). ii) A set of \textbf{Or-nodes} $V^{Or}$, in which each node branches to alternative decompositions (\eg, an indoor scene can be a bedroom or a living room), enabling the algorithm to reconfigure a scene. iii) A set of \textbf{Set nodes} $V^{Set}$, in which each node represents a nested And-Or relation: a set of Or-nodes serving as child branches are grouped by an And-node, and each child branch may include different numbers of objects.
	\item $V_{T} = V_T^r \cup V_T^a$. The terminal nodes consists of two subsets of nodes: regular nodes and address nodes. i) A \textbf{regular terminal node} $v \in V_T^r$ represents a spatial entity in a scene (\eg, an office chair in a bedroom) with attributes. In this paper, the attributes include internal attributes $A_{int}$ of object sizes $(w, l, h)$, external attributes $A_{ext}$ of object position $(x, y, z)$ and orientation ($x-y$ plane) $\theta$, and sampled human positions $A_h$. ii) To avoid excessively dense graphs, an \textbf{address terminal node} $v \in V_T^a$ is introduced to encode interactions that only occur in a certain context but are absent in all others~\cite{fridman2003mixed}. It is a pointer to regular terminal nodes, taking values in the set $V_T^r \cup \{\mbox{nil}\}$, representing supporting or grouping relations as shown in Figure~\ref{fig:representation}. 
\end{itemize}


\begin{figure}[t!]
	\begin{center}
		\includegraphics[width=\linewidth]{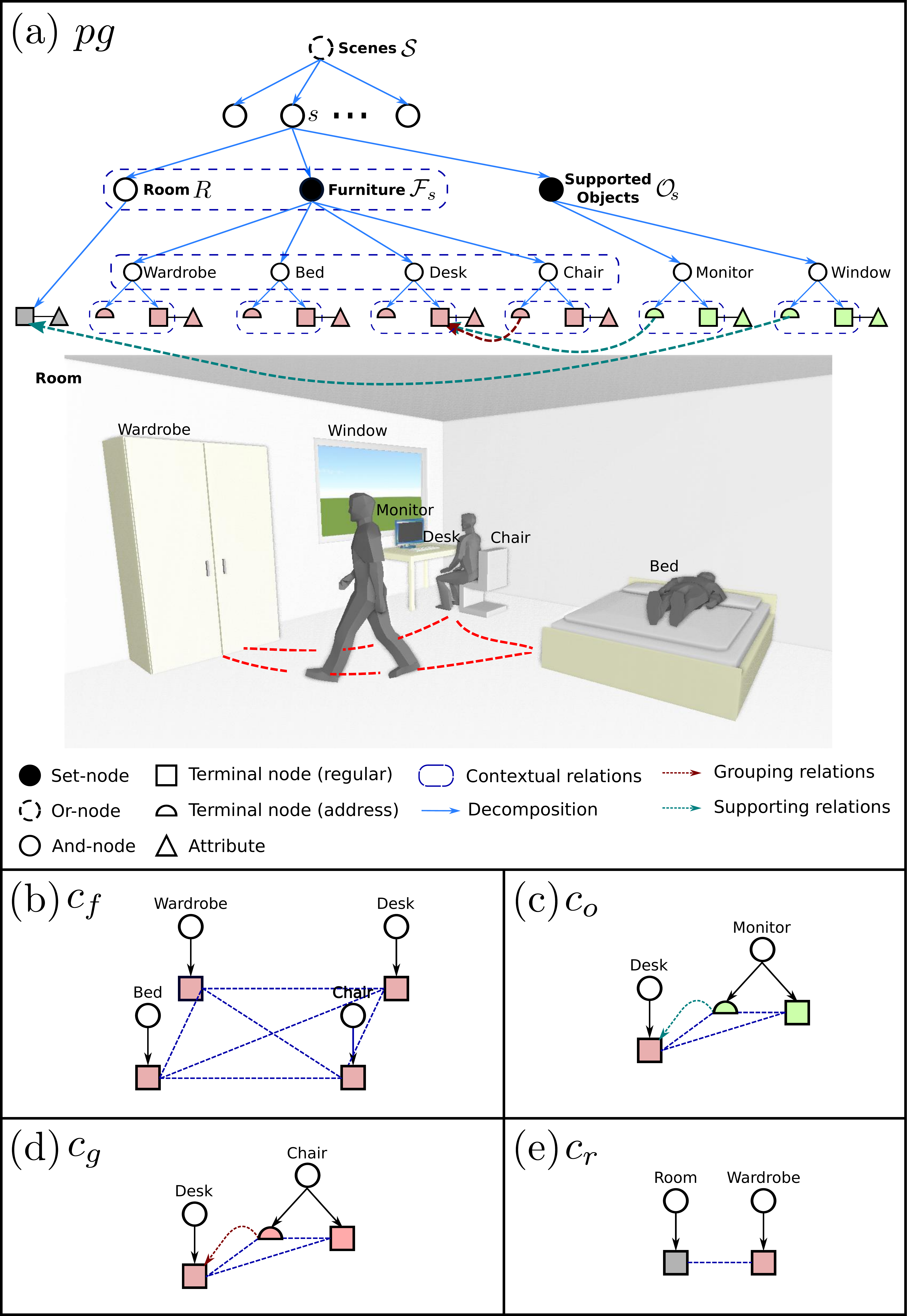}
	\end{center}
	\vspace{-10pt}
	\caption{(a) A simplified example of a parse graph of a bedroom. The terminal nodes of the parse graph form an MRF in the terminal layer. Cliques are formed by the contextual relations projected to the terminal layer. Examples of the four types of cliques are shown in (b)-(e), representing four different types of contextual relations.}
	\vspace{-10pt}
	\label{fig:parse_graph}
\end{figure}

\textbf{Contextual Relations} $E$ among nodes are represented by the horizontal links in S-AOG forming MRFs on the terminal nodes. To encode the contextual relations, we define different types of potential functions for different cliques.
The contextual relations $E = E_f \cup E_o \cup E_g \cup E_r$ are divided into four subsets: i) relations among furniture $E_f$; ii) relations between supported objects and their supporting objects $E_o$ (\eg, a monitor on a desk); iii) relations between objects of a functional pair $E_g$ (\eg, a chair and a desk); and iv) relations between furniture and the room $E_r$. Accordingly, the cliques formed in the terminal layer could also be divided into four subsets: $C = C_f \cup C_o \cup C_g \cup C_r$. Instead of directly capturing the object-object relations, we compute the potentials using affordances as a bridge to characterize the object-human-object relations.

A hierarchical parse tree $pt$ is an instantiation of the S-AOG by selecting a child node for the Or-nodes as well as determining the state of each child node for the Set-nodes. A parse graph $pg$ consists of a parse tree $pt$ and a number of contextual relations $E$ on the parse tree: $pg = (pt, E_{pt})$. Figure~\ref{fig:parse_graph} illustrates a simple example of a parse graph and four types of cliques formed in the terminal layer.

\section{Probabilistic Formulation of S-AOG}\label{sec:formulation}
A scene configuration is represented by a parse graph $pg$, including objects in the scene and associated attributes. The prior probability of $pg$ generated by an S-AOG parameterized by $\Theta$ is formulated as a Gibbs distribution:
{\small
\begin{align}
p(pg | \Theta) & = \frac{1}{Z} \exp \{ -\mathcal{E}(pg | \Theta) \} \\
& = \frac{1}{Z} \exp \{ -\mathcal{E}(pt | \Theta) - \mathcal{E}(E_{pt} | \Theta) \},
\end{align}
}where $\mathcal{E}(pg | \Theta)$ is the energy function of a parse graph, $\mathcal{E}(pt | \Theta)$ is the energy function of a parse tree, and $\mathcal{E}(E_{pt} | \Theta)$ is the energy term of the contextual relations. 

$\mathcal{E}(pt | \Theta)$ can be further decomposed into the energy functions of different types of non-terminal nodes, and the energy functions of internal attributes of both regular and address terminal nodes:
{\small
\begin{align}
\mathcal{E}(pt | \Theta)
& = \underbrace{
\sum_{\mathclap{v \in V^{Or}}} \mathcal{E}_{\Theta}^{Or}(v)
+ \sum_{\mathclap{v \in V^{Set}}} \mathcal{E}_{\Theta}^{Set}(v) }_\text{non-terminal nodes}
+ \underbrace{
\sum_{\mathclap{v \in V_{T}^{r}}} \mathcal{E}_{\Theta}^{A_{in}}(v) }_\text{terminal nodes},
\end{align}
}where the choice of the child node of an Or-node $v \in V^{Or}$ and the child branch of a Set-node $v \in V^{Set}$ follow different multinomial distributions. Since the And-nodes are deterministically expanded, we do not have an energy term for the And-nodes here. The internal attributes $A_{in}$ (size) of terminal nodes follows a non-parametric probability distribution learned by kernel density estimation.

$\mathcal{E}(E_{pt} | \Theta)$ combines the potentials of the four types of cliques formed in the terminal layer, integrating human attributes and external attributes of regular terminal nodes:
{\small
\begin{align}
p(E_{pt} | \Theta) & = \frac{1}{Z} \exp \{ -\mathcal{E}(E_{pt} | \Theta) \} \\
& = 
\prod_{\mathclap{c \in C_f}} \phi_f(c)
\prod_{\mathclap{c \in C_o}} \phi_o(c)
\prod_{\mathclap{c \in C_g}} \phi_g(c)
\prod_{\mathclap{c \in C_r}} \phi_r(c).
\end{align}
\label{eq:prob_clique}
}

\begin{figure}[t!]
	\centering
	\subfloat[Planned trajectories]{ \includegraphics[width=0.45\linewidth]{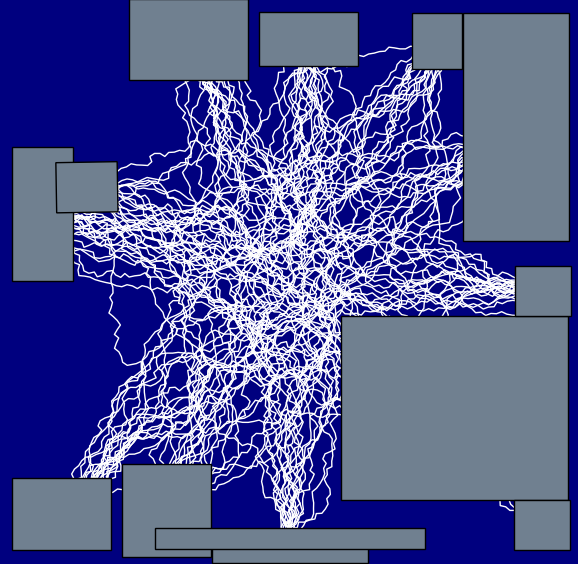} }
	\subfloat[Probability map]{ \includegraphics[width=0.45\linewidth]{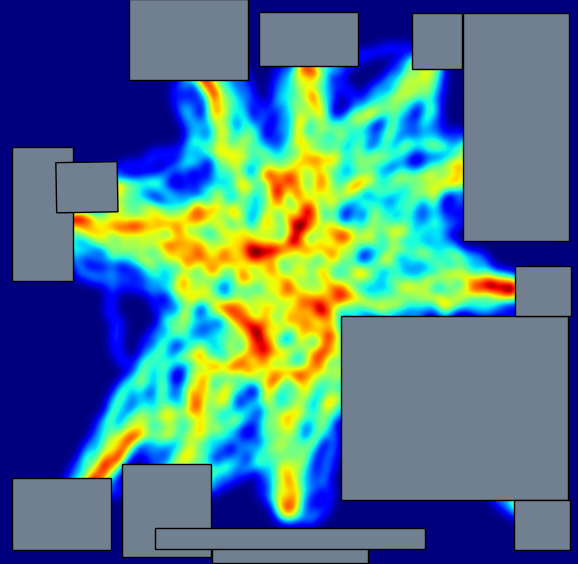} }
	\caption{Given a scene configuration, we use bi-directional RRT to plan from every piece of furniture to another, generating a human activity probability map.}
	\label{fig:planning}
\end{figure}

\noindent\textbf{Human Centric Potential Functions:}\vspace{3pt}
\begin{itemize}[wide,leftmargin=0cm,noitemsep,nolistsep]
\item Potential function $\phi_f(c)$ is defined on relations between furniture (Figure~\ref{fig:parse_graph}(b)). 
The clique $c = \{f_i\} \in C_f$ includes all the terminal nodes representing furniture:
{\small
\begin{align}
\phi_f(c) = \frac{1}{Z} \exp \{ - \lambda_f \cdot \langle \sum_{f_i \neq f_j} l_{\textup{col}}(f_i, f_j), l_{\textup{ent}}(c) \rangle \},
\end{align}
}where $\lambda_f$ is a weight vector, $<\cdot, \cdot>$ denotes a vector, and the cost function $l_{\textup{col}}(f_i, f_j)$ is the overlapping volume of the two pieces of furniture, serving as the penalty of collision. 
The cost function $l_{\textup{ent}}(c) = -H(\Gamma) = \Sigma_i p(\gamma_i)\log p(\gamma_i) $ yields better utility of the room space by sampling human trajectories, where $\Gamma$ is the set of planned trajectories in the room, and $H(\Gamma)$ is the entropy. The trajectory probability map is first obtained by planning a trajectory $\gamma_i$ from the center of every piece of furniture to another one using bi-directional rapidly-exploring random tree (RRT)~\cite{lavalle1998rapidly}, which forms a heatmap. The entropy is computed from the heatmap as shown in Figure~\ref{fig:planning}.

\item Potential function $\phi_o(c)$ is defined on relations between a supported object and the supporting furniture (Figure~\ref{fig:parse_graph}(c)).
A clique $c = \{f, a, o\} \in C_o$ includes a supported object terminal node $o$, the address node $a$ connected to the object, and the furniture terminal node $f$ pointed by $a$:
{\small
\begin{align}
\phi_o(c) = \frac{1}{Z} \exp \{ - \lambda_o \cdot \langle l_{\textup{hum}}(f, o) , l_{\textup{add}}(a) \rangle \},
\end{align}
}where the cost function $l_{\textup{hum}}(f, o)$ defines the human usability cost---a favorable human position should enable an agent to access or use both the furniture and the object. To compute the usability cost, human positions $h_i^o$ are first sampled based on position, orientation, and the affordance map of the supported object. Given a piece of furniture, the probability of the human positions is then computed by:
{\small
\begin{align}
l_{\textup{hum}}(f, o) = \max_i p(h_i^o | f).
\end{align}
}The cost function $l_{\textup{add}}(a)$ is the negative log probability of an address node $v \in V_{T}^{a}$, treated as a certain regular terminal node, following a multinomial distribution.

\item Potential function $\phi_g(c)$ is defined on functional grouping relations between furniture (Figure~\ref{fig:parse_graph}(d)). A clique $c = \{f_i, a, f_j\} \in C_g$ consists of terminal nodes of a core functional furniture $f_i$, pointed by the address node $a$ of an associated furniture $f_j$. The grouping relation potential is defined similarly to the supporting relation potential
{\small
\begin{align}
\phi_g(c) = \frac{1}{Z} \exp \{ - \lambda_c \cdot \langle l_{\textup{hum}}(f_i, f_j), l_{\textup{add}}(a) \rangle \}.
\end{align}
}
\end{itemize}

\noindent\textbf{Other Potential Functions:}\vspace{3pt}
\begin{itemize}[wide,leftmargin=0cm,noitemsep,nolistsep]
\item Potential function $\phi_r(c)$ is defined on relations between the room and furniture (Figure~\ref{fig:parse_graph}(e)).
A clique $c = \{ f, r \} \in C_r$ includes a terminal node $f$ and $r$ representing a piece of furniture and a room, respectively. The potential is defined as
{\small
\begin{align}
\phi_r(c) = 
\frac{1}{Z} \exp \{- \lambda_r \cdot \langle l_{\textup{dis}}(f, r), l_{\textup{ori}}(f, r) \rangle \},
\end{align}
}where the distance cost function is defined as $l_{\textup{dis}}(f, r) = -\log p(d | \Theta)$, in which $d \sim \ln \mathcal{N}(\mu, \sigma^2)$ is the distance between the furniture and the nearest wall modeled by a log normal distribution. The orientation cost function is defined as $l_{\textup{ori}}(f, r) = -\log p(\theta | \Theta)$, where $\theta \sim p(\mu, \kappa) = \frac{e^{\kappa \cos(x-\mu)}}{2 \pi I_0(\kappa)}$ is the relative orientation between the model and the nearest wall modeled by a von Mises distribution.
\end{itemize}

\section{Learning S-AOG}\label{sec:learning}
We use the SUNCG dataset~\cite{song2016ssc} as training data. It contains over 45K different scenes with manually created realistic room and furniture layouts. We collect the statistics of room types, room sizes, furniture occurrences, furniture sizes, relative distances, orientations between furniture and walls, furniture affordance, grouping occurrences, and supporting relations. 
The parameters $\Theta$ of the probability model $P$ can be learned in a supervised way by maximum likelihood estimation (MLE).

\textbf{Weights of Loss Functions:} Recall that the probability distribution of cliques formed in the terminal layer is
{\small
\begin{align}
p(E_\mathit{pt} | \Theta) & = \frac{1}{Z} \exp \{ -\mathcal{E}(E_\mathit{pt} | \Theta) \} \\
& = \frac{1}{Z} \exp \{ - \langle \lambda, l(E_\mathit{pt}) \rangle \},
\end{align}}where $\lambda$ is the weight vector and $l(E_\mathit{pt})$ is the loss vector given by four different types of potential functions.

To learn the weight vector, the standard MLE maximizes the average log-likelihood:
{\small
\begin{align}
\mathcal{L}(E_\mathit{pt} | \Theta) & = - \frac{1}{N} \sum_{n=1}^{N} \langle \lambda, l(E_\mathit{pt_n}) \rangle - \log Z.
\end{align}}

This is usually maximized by following the gradient:
{\small
\begin{align}
	& \frac{\partial \mathcal{L}(E_\mathit{pt} | \Theta)}{\partial \lambda}  = - \frac{1}{N} \sum_{n=1}^{N} l(E_\mathit{pt_n}) -
\frac{\partial \log Z}{\partial \lambda} \\
& = - \frac{1}{N} \sum_{n=1}^{N} l(E_\mathit{pt_n}) -
\frac{\partial \log \sum_{pt}{\exp\{- \langle \lambda, l(E_\mathit{pt}) \rangle\}}}{\partial \lambda} \\
& = - \frac{1}{N} \sum_{n=1}^{N} l(E_\mathit{pt_n}) + 
\sum_{pt} \frac{1}{Z} \exp \{ - \langle \lambda, l(E_\mathit{pt}) \rangle \} l(E_\mathit{pt}) \\
& = - \frac{1}{N} \sum_{n=1}^{N} l(E_\mathit{pt_n}) + 
\frac{1}{\widetilde{N}} \sum_{\widetilde{n}=1}^{\widetilde{N}} l(E_\mathit{pt_{\widetilde{n}}}),
\end{align}}where $\{E_\mathit{pt_{\widetilde{n}}}\}_{\widetilde{n}=1, \cdots, \widetilde{N}}$ is the set of synthesized examples from the current model. 

It is usually computationally infeasible to sample a Markov chain that burns into an \textit{equilibrium distribution} at every iteration of gradient ascent. Hence, instead of waiting for the Markov chain to converge, we adopt the contrastive divergence (CD) learning that follows the gradient of difference of two divergences~\cite{hinton2002training}
{\small
\begin{align}
\mbox{CD}_{\widetilde{N}} = \mbox{KL}(p_0 || p_{\infty}) - \mbox{KL}(p_{\widetilde{n}} || p_{\infty}),
\end{align}}where $\mbox{KL}(p_0 || p_{\infty})$ is the Kullback-Leibler divergence between the data distribution $p_0$ and the model distribution $p_{\infty}$, and $p_{\widetilde{n}}$ is the distribution obtained by a Markov chain started at the data distribution and run for a small number $\widetilde{n}$ of steps. In this paper, we set $\widetilde{n}=1$.

Contrastive divergence learning has been applied effectively to addressing various problems; one of the most notable work is in the context of Restricted Boltzmann Machines~\cite{hinton2006reducing}. Both theoretical and empirical evidences shows its efficiency while keeping bias typically very small~\cite{carreira2005contrastive}. The gradient of the contrastive divergence is given by
{\small
\begin{align}
\frac{\partial \mbox{CD}_{\widetilde{N}}}{\partial \lambda} = & \frac{1}{N} \sum_{n=1}^{N} l(E_\mathit{pt_n}) -
\frac{1}{\widetilde{N}} \sum_{\widetilde{n}=1}^{\widetilde{N}} l(E_\mathit{pt_{\widetilde{n}}}) \\\notag
& - \frac{\partial p_{\widetilde{n}}}{\partial \lambda} \frac{\partial \mbox{KL} (p_{\widetilde{n}} || p_{\infty}) }{\partial p_{\widetilde{n}}}.
\end{align}}Extensive simulations~\cite{hinton2002training} showed that the third term can be safely ignored since it is small and seldom opposes the resultant of the other two terms.

Finally, the weight vector is learned by gradient descent computed by generating a small number $\widetilde{N}$ of examples from the Markov chain
{\small
\begin{align}
\lambda_{t+1} & = \lambda_{t} - \eta_{t} \frac{\partial \mbox{CD}_{\widetilde{N}}}{\partial \lambda} \\
& = \lambda_{t} + \eta_{t} \left(
\frac{1}{\widetilde{N}} \sum_{\widetilde{n}=1}^{\widetilde{N}} l(E_\mathit{pt_{\widetilde{n}}}) -
\frac{1}{N} \sum_{n=1}^{N} l(E_\mathit{pt_n}) \right).
\end{align}}

\textbf{Branching Probabilities:} The MLE of the branch probabilities $\rho_i$ of Or-nodes, Set-nodes and address terminal nodes is simply the frequency of each alternative choice~\cite{zhu2007stochastic}: $\rho_i = \#(v \rightarrow u_i)/\sum\limits_{j=1}^{n(v)}\#(v \rightarrow u_j)$.

\begin{figure*}[t!]
    \centering
\begin{tabular}[c]{@{\hskip-1.4em}c@{\hskip-1em}c@{\hskip-1em}c@{\hskip-1em}c@{\hskip-1em}c@{\hskip-1em}}
\subfloat[desk]{ \includegraphics[width=0.16\linewidth]{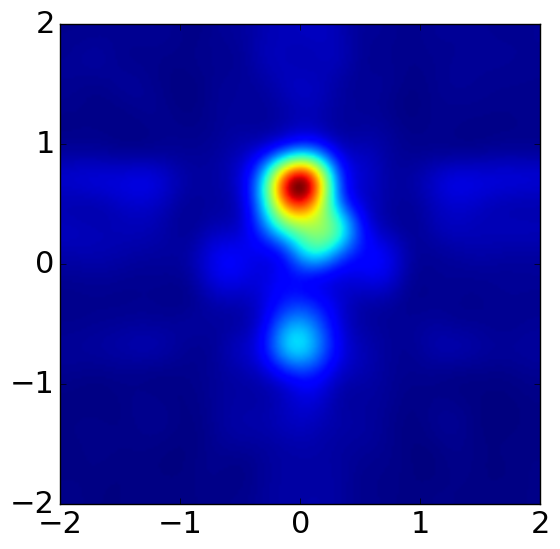} }
\subfloat[coffee table]{ \includegraphics[width=0.16\linewidth]{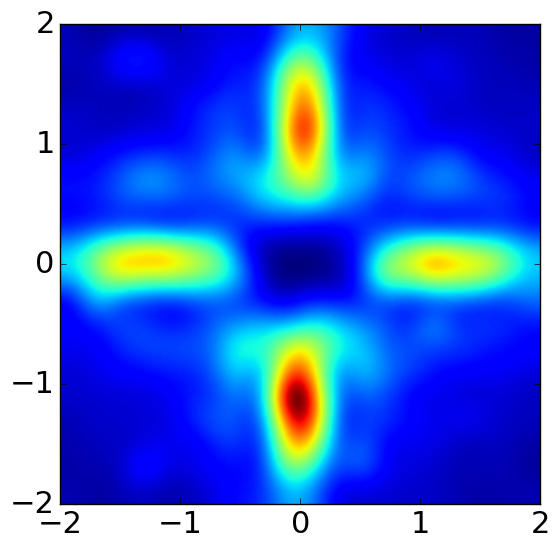} }
\subfloat[dining table]{ \includegraphics[width=0.16\linewidth]{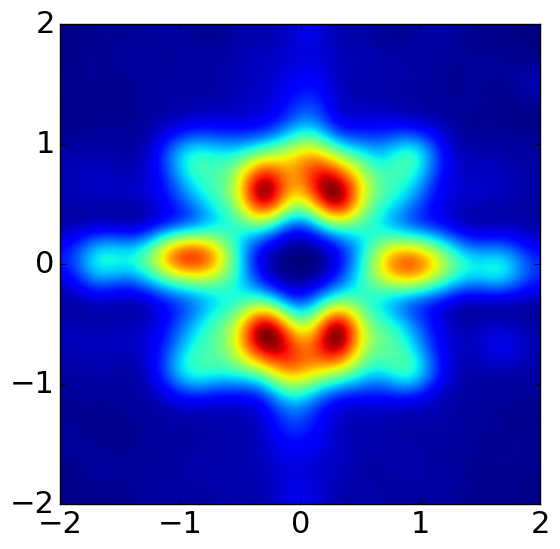} }
\subfloat[books]{ \includegraphics[width=0.16\linewidth]{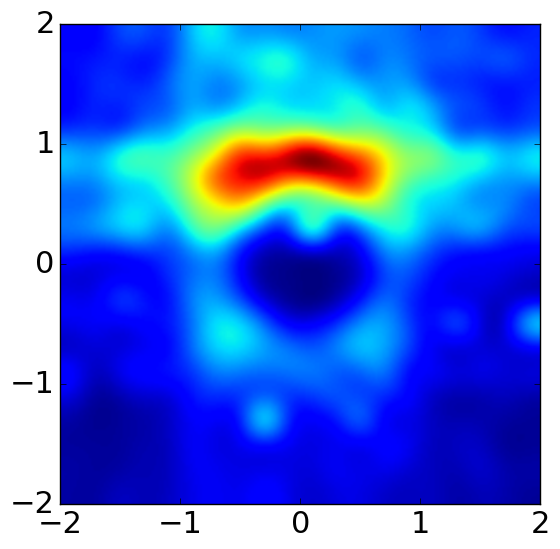} }
\subfloat[laptop]{ \includegraphics[width=0.16\linewidth]{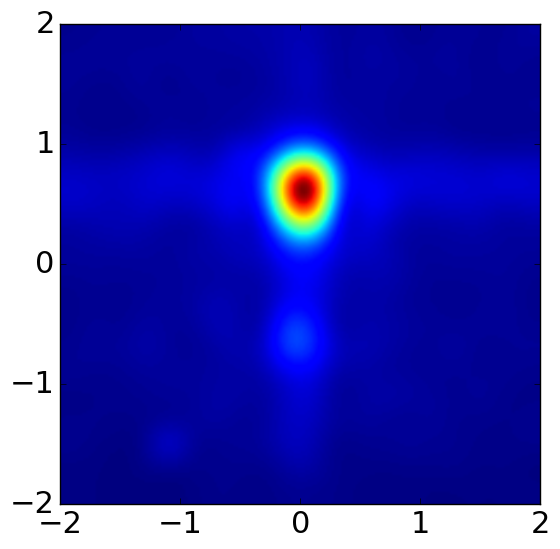} }
\subfloat[nightstand]{ \includegraphics[width=0.16\linewidth]{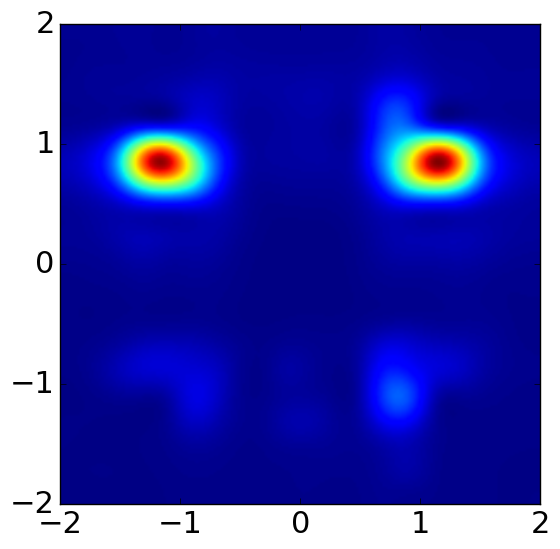} }  \vspace{-10pt} \\
\subfloat[fruit bowl]{ \includegraphics[width=0.16\linewidth]{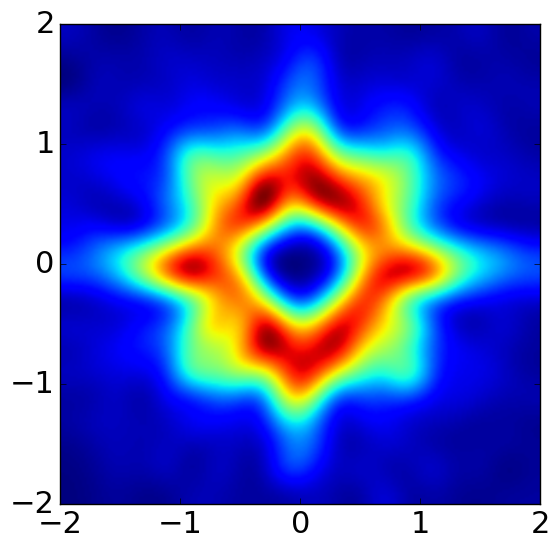} }
\subfloat[vase]{ \includegraphics[width=0.16\linewidth]{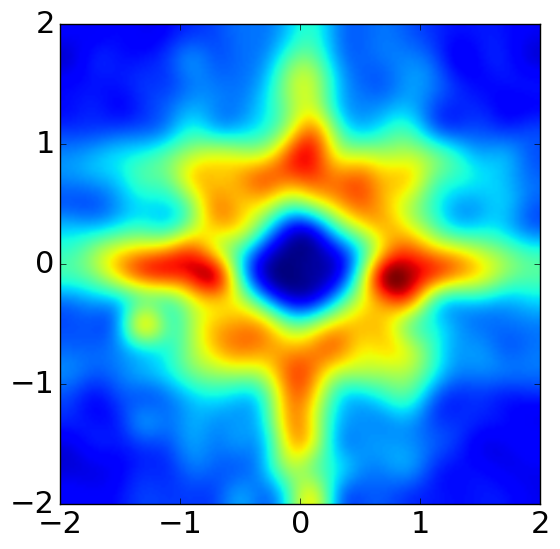} }
\subfloat[floor lamp]{ \includegraphics[width=0.16\linewidth]{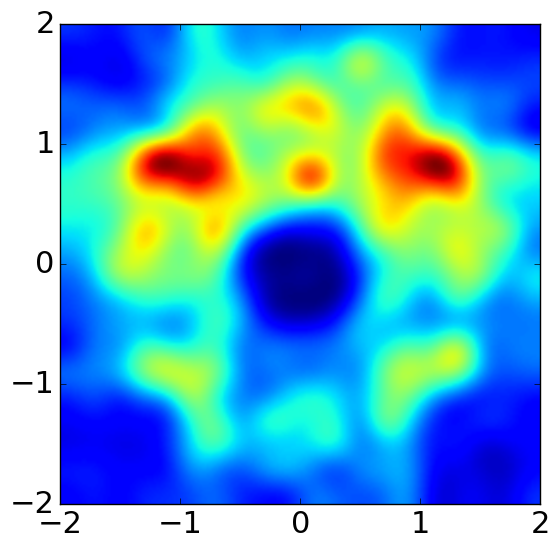} }
\subfloat[wall lamp]{ \includegraphics[width=0.16\linewidth]{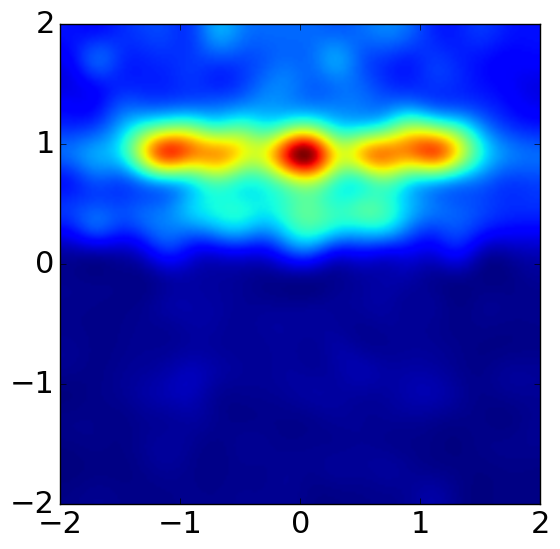} }
\subfloat[fireplace]{ \includegraphics[width=0.16\linewidth]{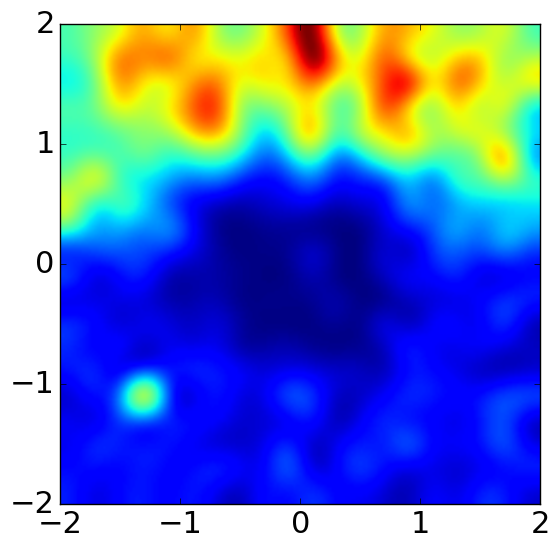} }
\subfloat[ceiling fan]{ \includegraphics[width=0.16\linewidth]{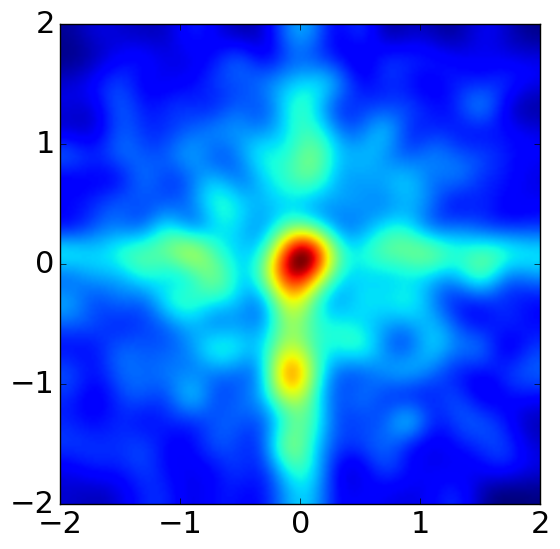} }
\end{tabular}
    \caption{Examples of the learned affordance maps. Given the object positioned in the center facing upwards, \ie, coordinate of $(0, 0)$ facing direction $(0, 1)$, the maps show the distributions of human positions. The affordance maps accurately capture the subtle differences among desks, coffee tables, and dining tables. Some objects are orientation sensitive, \eg, books, laptops, and night stands, while some are orientation invariant, \eg, fruit bowls and vases.}
    \label{fig:affordance}
\end{figure*}

\begin{figure*}[t!]
	\begin{center}
		\includegraphics[width=\linewidth]{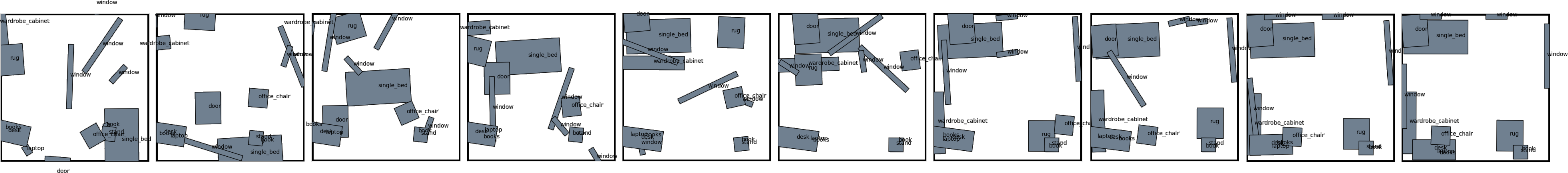}
	\end{center}
	\vspace{-10pt}
	\caption{MCMC sampling process (from left to right) of scene configurations with simulated annealing.}
	\label{fig:mcmc}
\end{figure*}

\textbf{Grouping Relations:} The grouping relations are hand-defined (\ie, nightstands are associated with beds, chairs are associated with desks and tables). The probability of occurrence is learned as a multinomial distribution, and the supporting relations are automatically extracted from SUNCG.

\textbf{Room Size and Object Sizes:} The distribution of the room size and object size among all the furniture and supported objects is learned as a non-parametric distribution. We first extract the size information from the 3D models inside SUNCG dataset, and then fit a non-parametric distribution using kernel density estimation. The distances and relative orientations of the furniture and objects to the nearest wall are computed and fitted into a log normal and a mixture of von Mises distributions, respectively.

\textbf{Affordances:} We learn the affordance maps of all the furniture and supported objects by computing the heatmap of possible human positions. These position include annotated humans, and we assume that the center of chairs, sofas, and beds are positions that humans often visit. By accumulating the relative positions, we get reasonable affordance maps as non-parametric distributions as shown in Figure~\ref{fig:affordance}.

\section{Synthesizing Scene Configurations}\label{sec:synthesis}
Synthesizing scene configurations is accomplished by sampling a parse graph $pg$ from the prior probability $p(pg | \Theta)$ defined by the S-AOG. The structure of a parse tree $pt$ (\ie, the selection of Or-nodes and child branches of Set-nodes) and the internal attributes (sizes) of objects can be easily sampled from the closed-form distributions or non-parametric distributions. However, the external attributes (positions and orientations) of objects are constrained by multiple potential functions, hence they are too complicated to be directly sampled from. Here, we utilize a Markov chain Monte Carlo (MCMC) sampler to draw a typical state in the distribution. The process of each sampling can be divided into two major steps:
\begin{enumerate}[wide,leftmargin=0cm,noitemsep,nolistsep]
    \item Directly sample the structure of $pt$ and internal attributes $A_{in}$: (i) sample the child node for the Or-nodes; (ii) determine the state of each child branch of the Set-nodes; and (iii) for each regular terminal node, sample the sizes and human positions from learned distributions.
    \item Use an MCMC scheme to sample the values of address nodes $V^a$ and external attributes $A_{ex}$ by making proposal moves. A sample will be chosen after the Markov chain converges.
\end{enumerate}

We design two simple types of Markov chain dynamics which are used at random with probabilities $q_i, i=1, 2$ to make proposal moves:
\begin{itemize}[wide,leftmargin=0cm,noitemsep,nolistsep]
\item Dynamics $q_1$: translation of objects. This dynamic chooses a regular terminal node, and samples a new position based on the current position $x$: $x \rightarrow x + \delta x$, where $\delta x$ follows a bivariate normal distribution.

\item Dynamics $q_2$: rotation of objects. This dynamic chooses a regular terminal node, and samples a new orientation based on the current orientation of the object: $\theta \rightarrow \theta + \delta \theta$, where $\delta \theta$ follows a normal distribution.
\end{itemize}

Adopting the Metropolis-Hastings algorithm, the proposed new parse graph $pg'$ is accepted according to the following acceptance probability:
{\small
\begin{align}
 \alpha(pg' | pg, \Theta) &= \min(1, \frac{p(pg'|\Theta) p(pg | pg')}{p(pg|\Theta) p(pg' | pg)}) \\
 &= \min(1, \exp(\mathcal{E}(pg|\Theta) - \mathcal{E}(pg'|\Theta))),
\end{align}
}where the proposal probability rate is canceled since the proposal moves are symmetric in probability. A simulated annealing scheme is adopted to obtain samples with high probability as shown in Figure~\ref{fig:mcmc}.

\begin{figure*}[t!]
\begin{center}
	\vspace{-10pt}
    \begin{tabular}[c]{@{\hskip-1.4em}c@{\hskip-1em}c@{\hskip-1em}c@{\hskip-1em}c@{\hskip-1em}c@{\hskip-1em}}
        \captionsetup[subfigure]{aboveskip=-6pt,belowskip=-6pt}
        \subfloat[bathroom]{
        \begin{tabular}[b]{c}
            \includegraphics[width=0.195\linewidth]{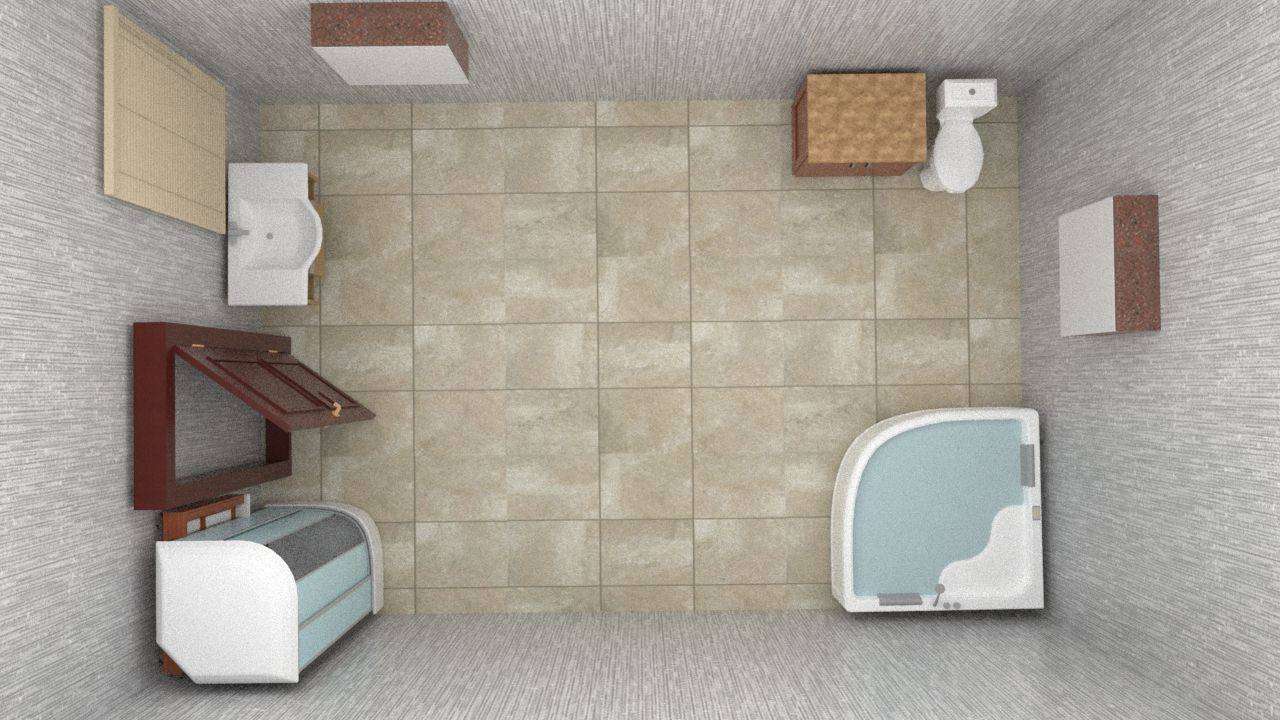} \\
            \includegraphics[width=0.195\linewidth]{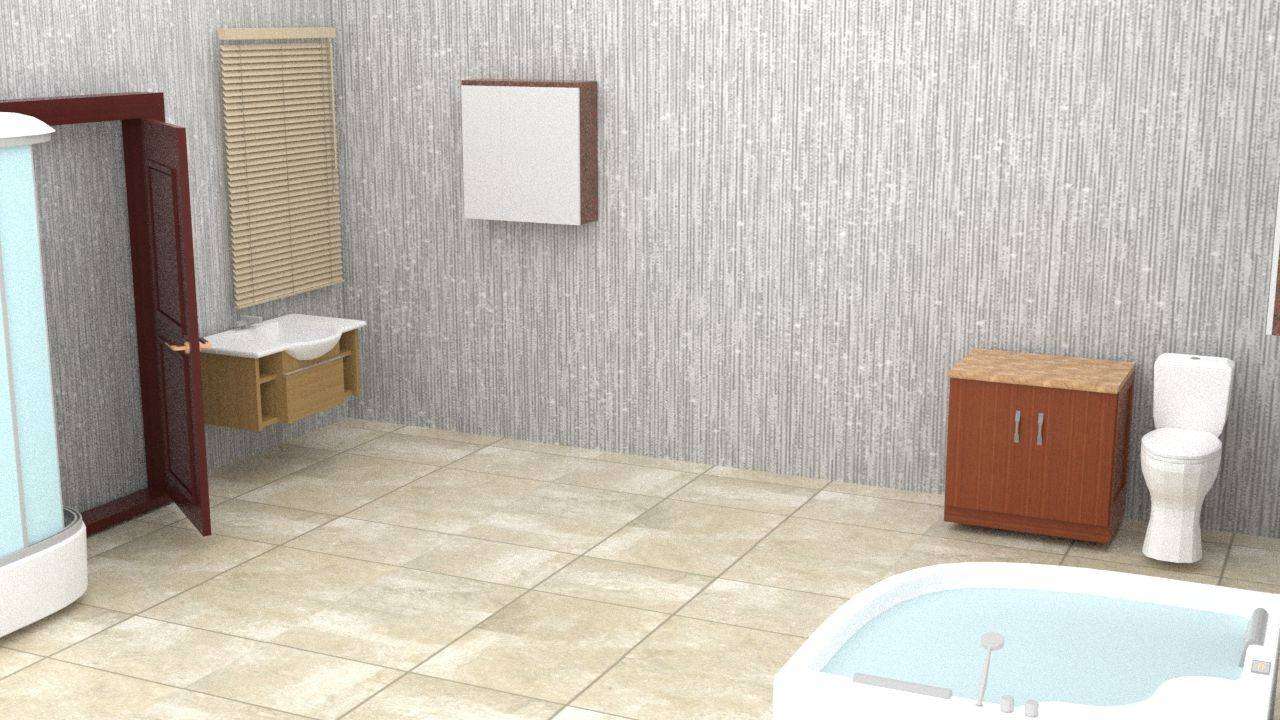} \\
            \includegraphics[width=0.195\linewidth]{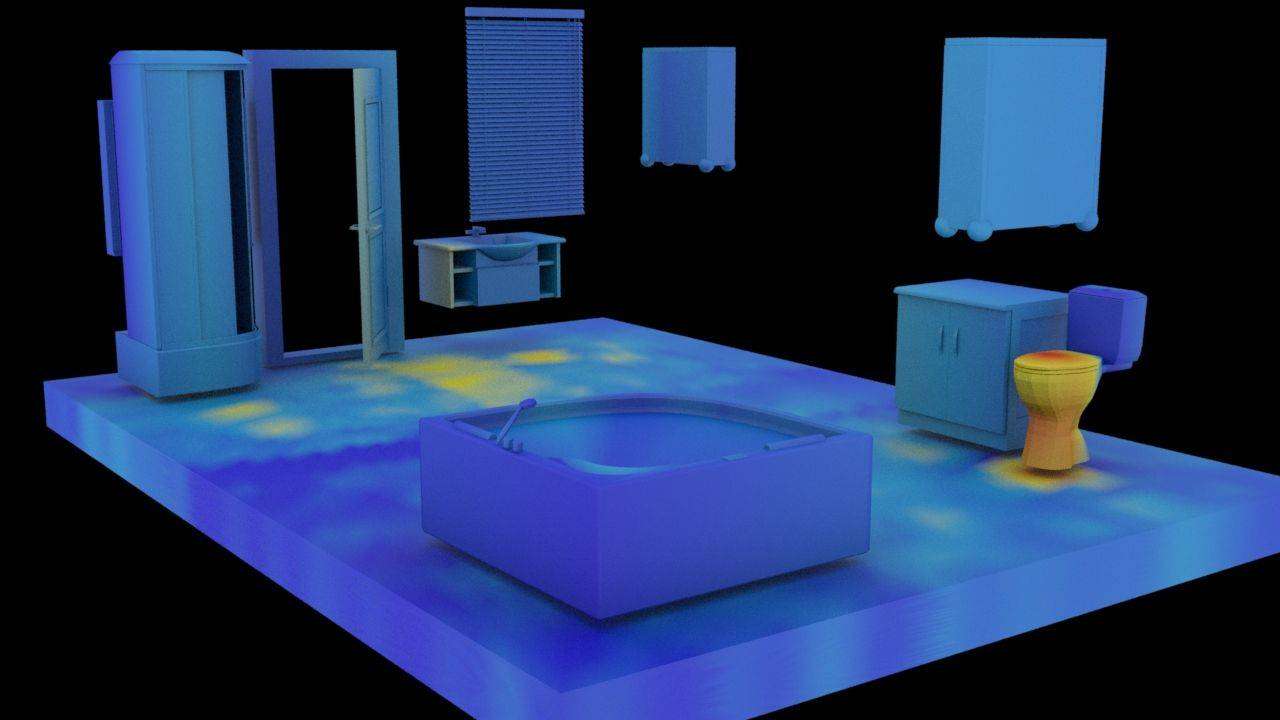}
        \end{tabular}}\hfill
        &
        \subfloat[bedroom]{
        \begin{tabular}[b]{c}
            \includegraphics[width=0.195\linewidth]{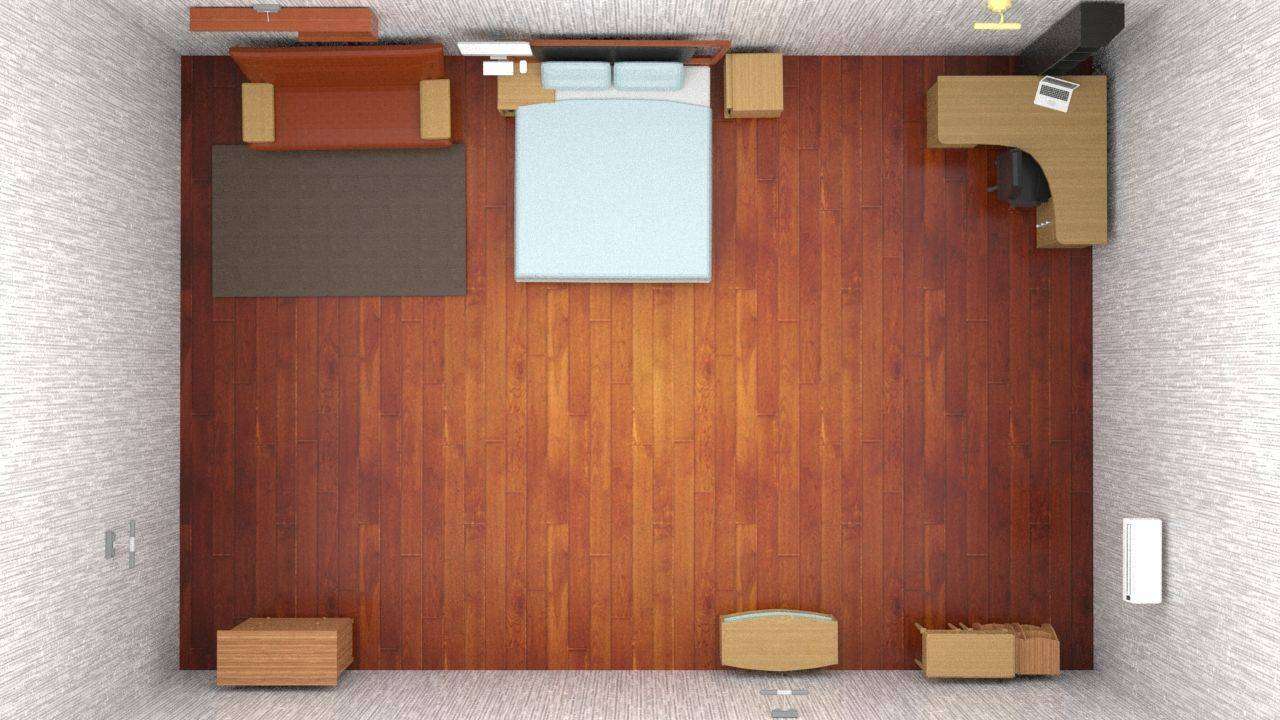} \\
            \includegraphics[width=0.195\linewidth]{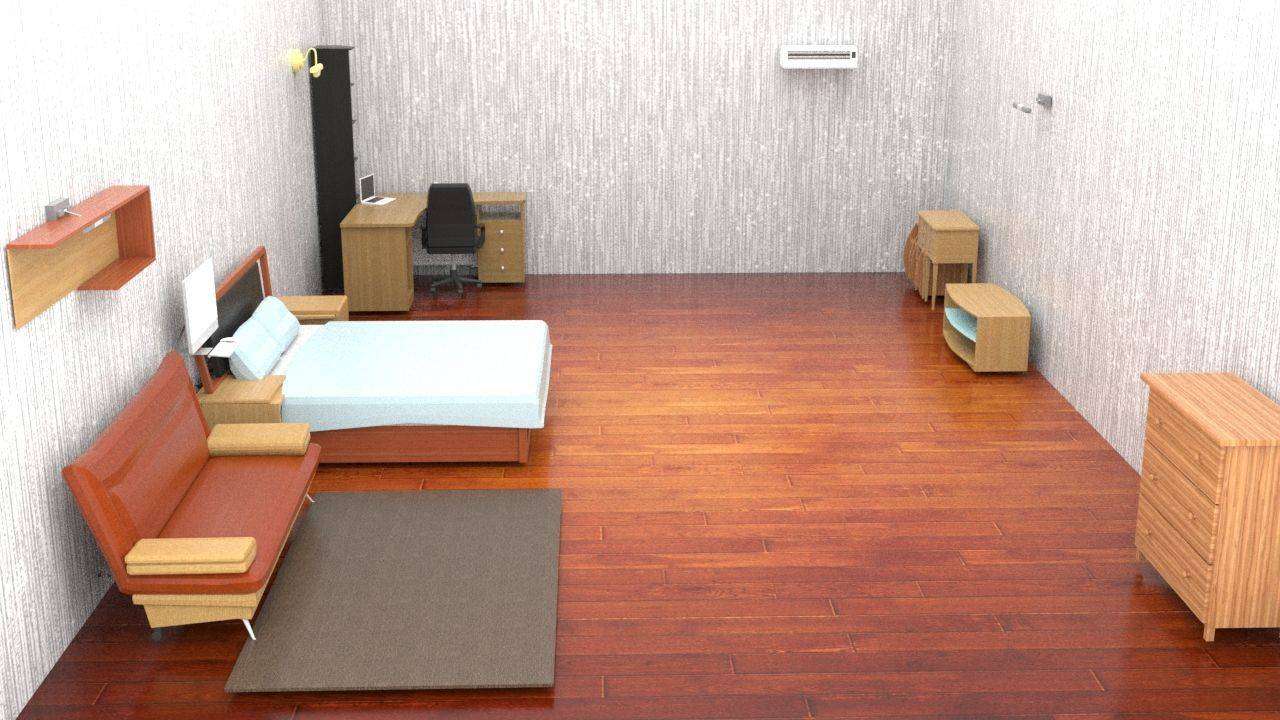} \\
            \includegraphics[width=0.195\linewidth]{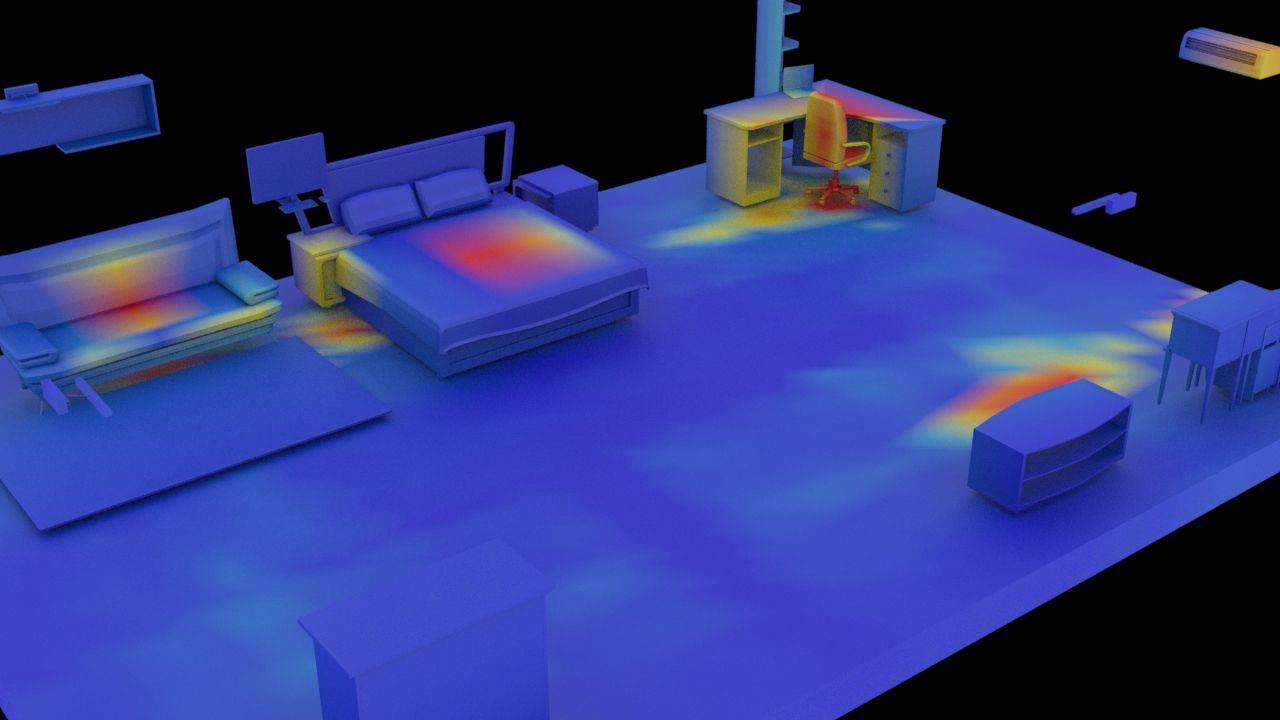}
        \end{tabular}}\hfill
        &
        \subfloat[dining room]{
        \begin{tabular}[b]{c}
            \includegraphics[width=0.195\linewidth]{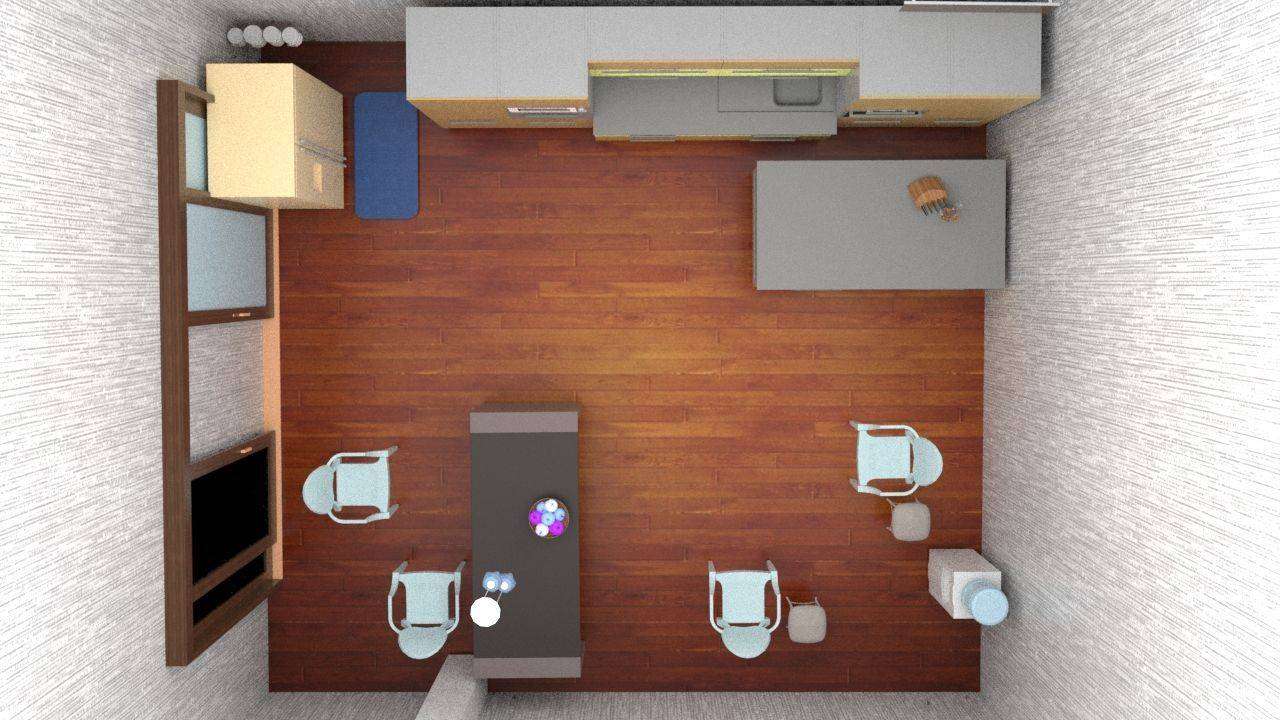} \\
            \includegraphics[width=0.195\linewidth]{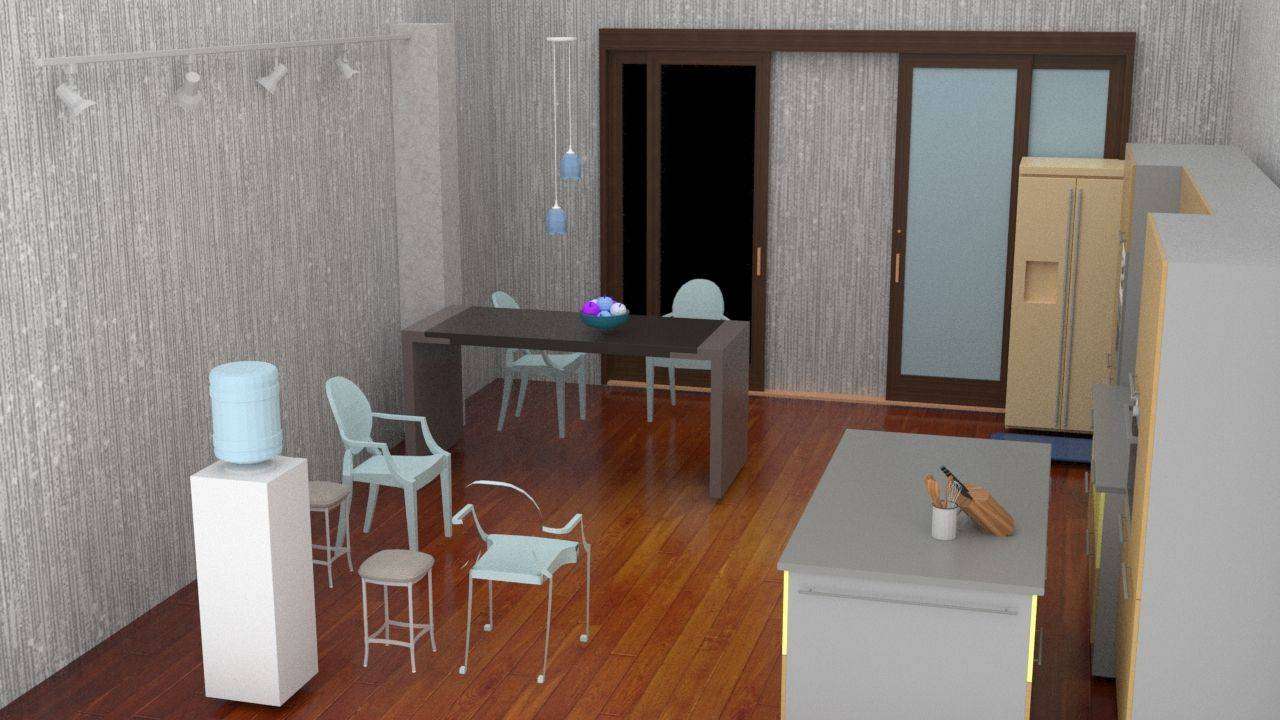} \\
            \includegraphics[width=0.195\linewidth]{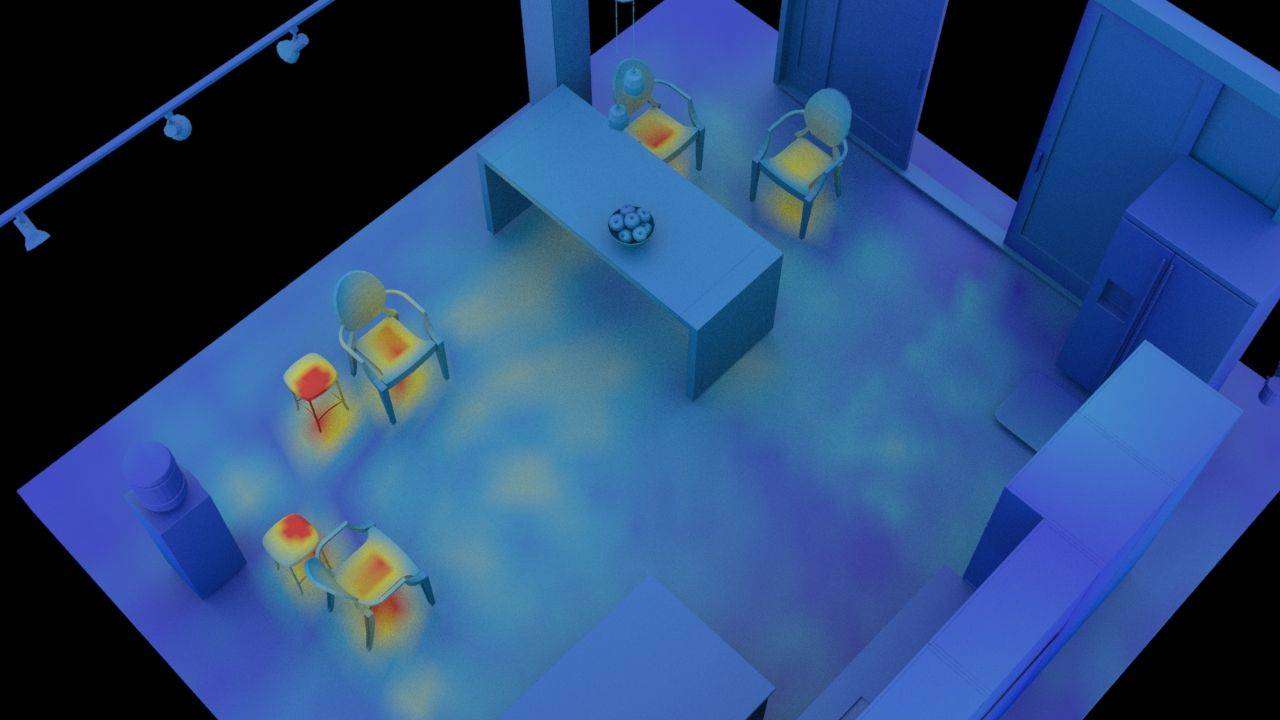}
        \end{tabular}}\hfill
        &
        \subfloat[garage]{
        \begin{tabular}[b]{c}
            \includegraphics[width=0.195\linewidth]{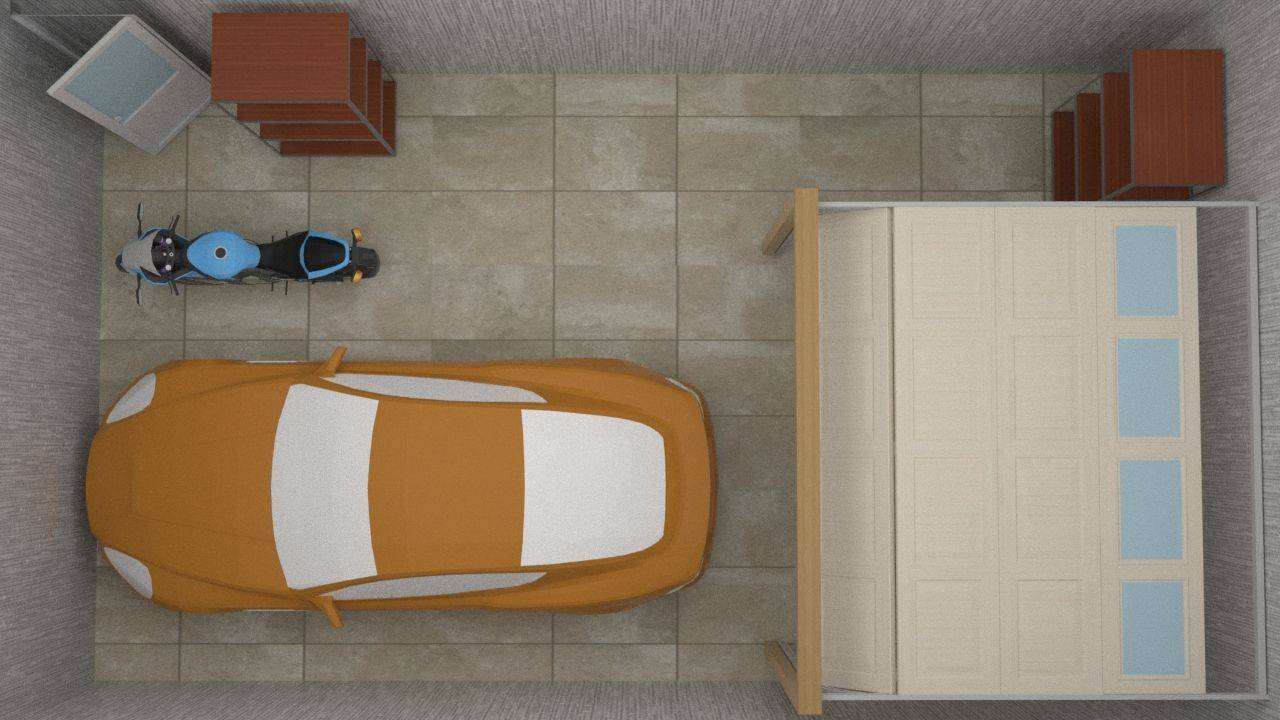} \\
            \includegraphics[width=0.195\linewidth]{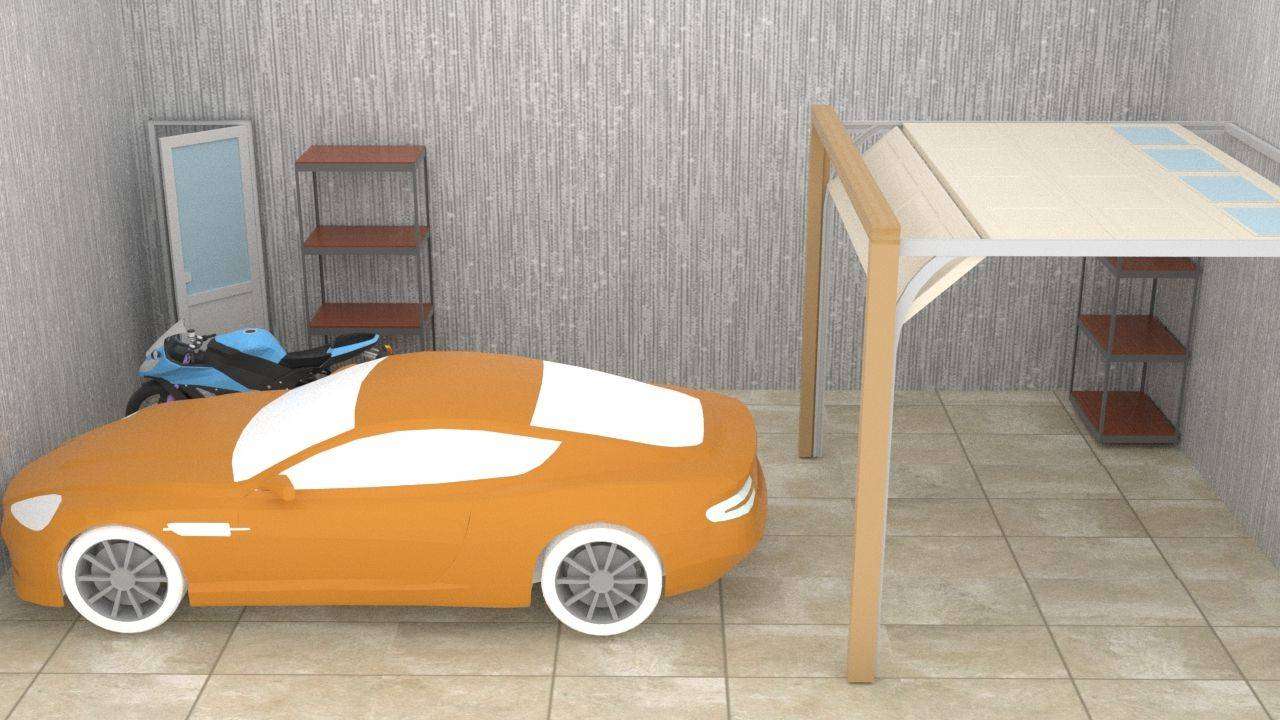} \\
            \includegraphics[width=0.195\linewidth]{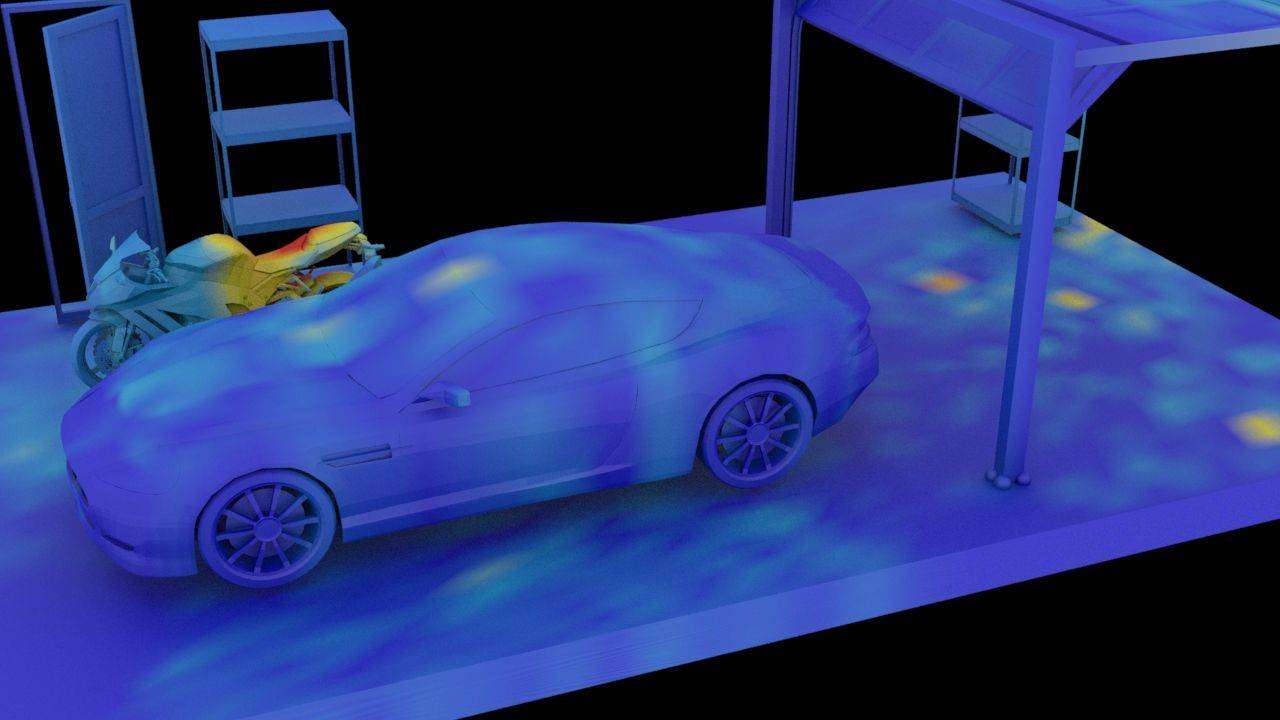}
        \end{tabular}}\hfill
        &
        \subfloat[guest room]{
        \begin{tabular}[b]{c}
            \includegraphics[width=0.195\linewidth]{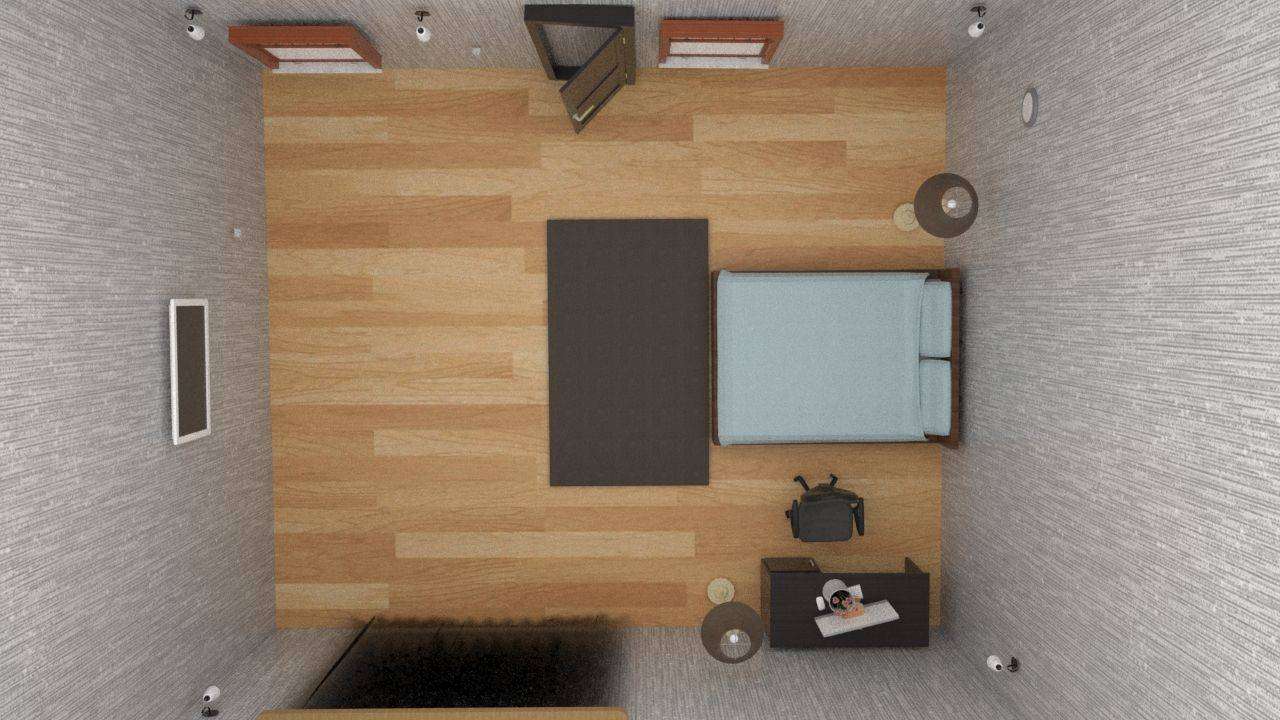} \\
            \includegraphics[width=0.195\linewidth]{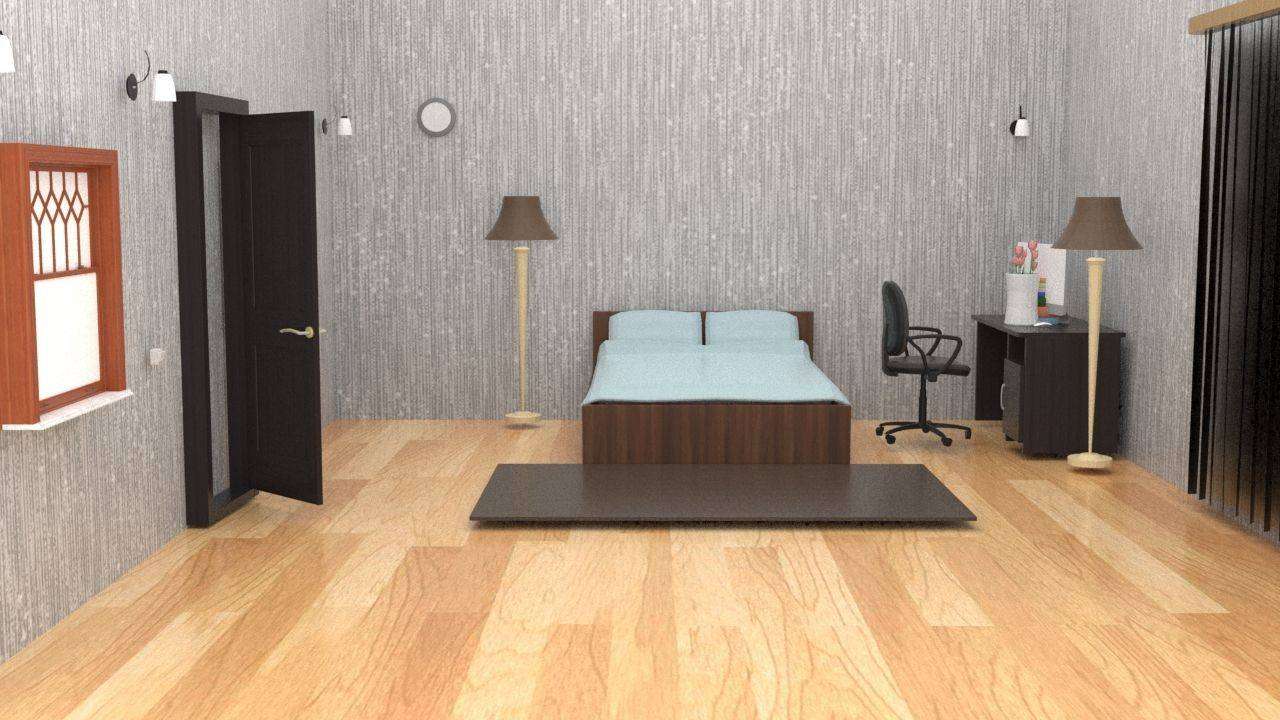} \\
            \includegraphics[width=0.195\linewidth]{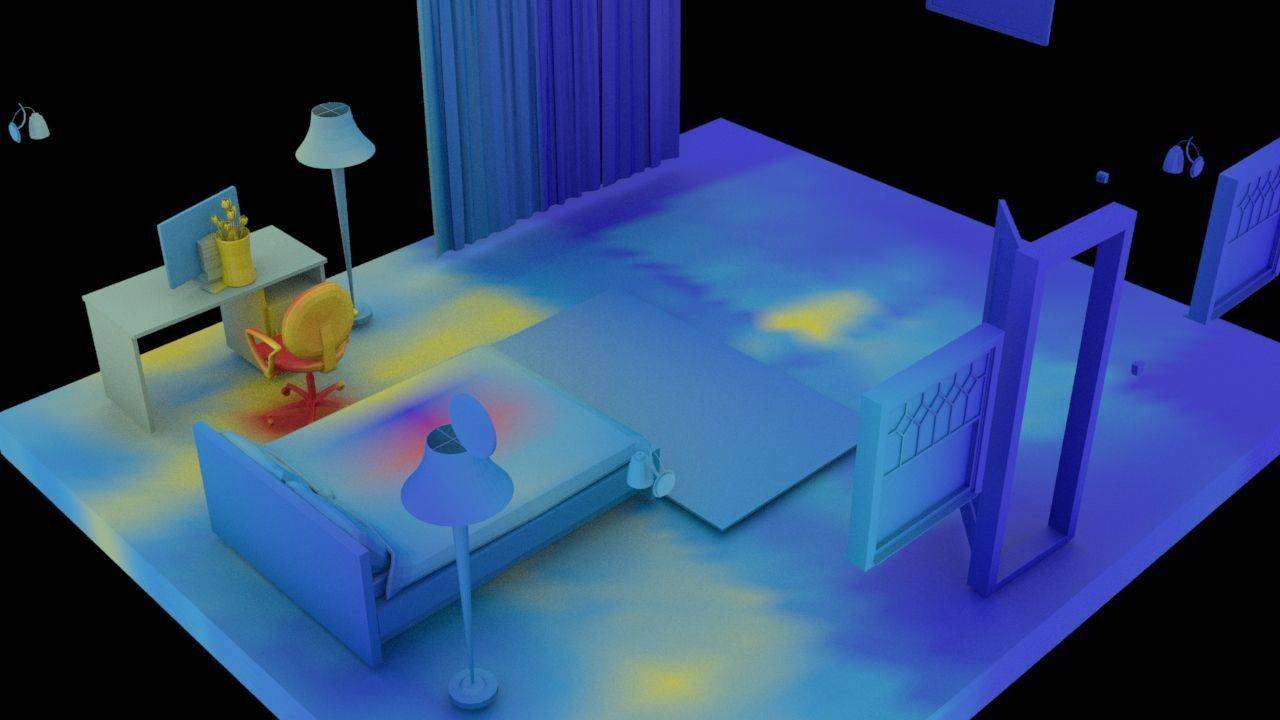}
        \end{tabular}}\hfill \\ [-1.5ex]
        \subfloat[gym]{
        \begin{tabular}[b]{c}
            \includegraphics[width=0.195\linewidth]{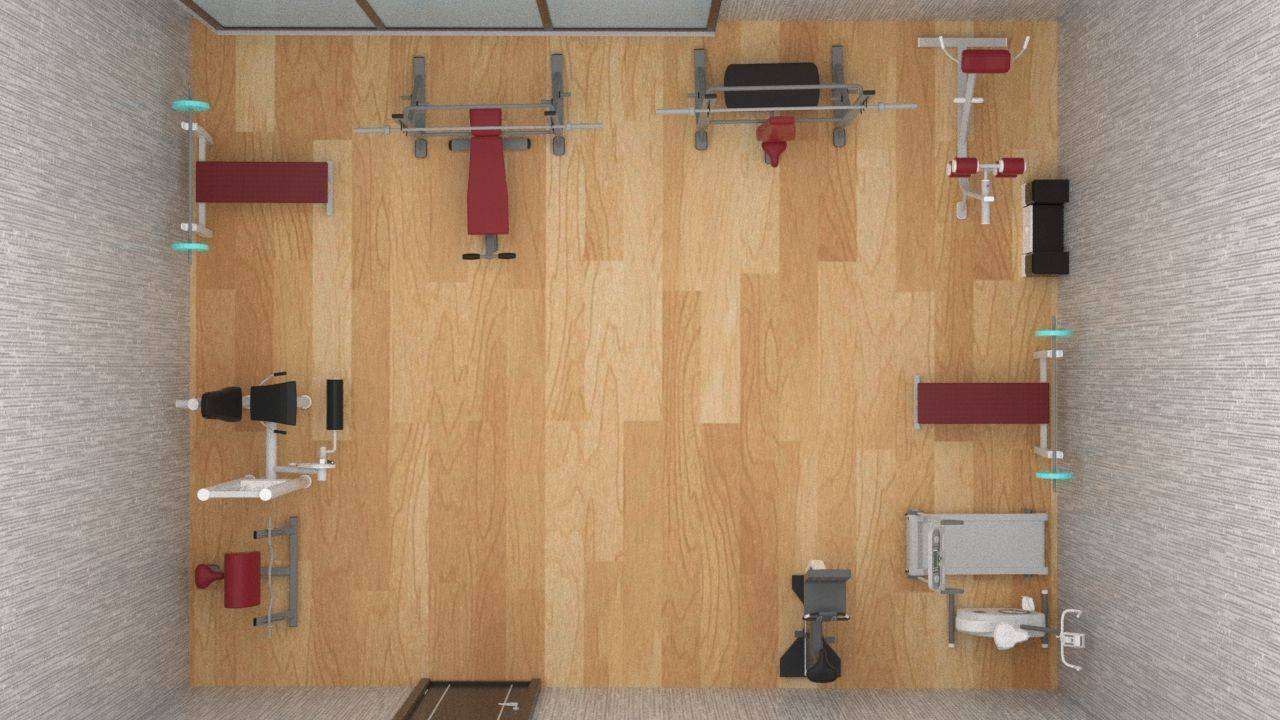} \\
            \includegraphics[width=0.195\linewidth]{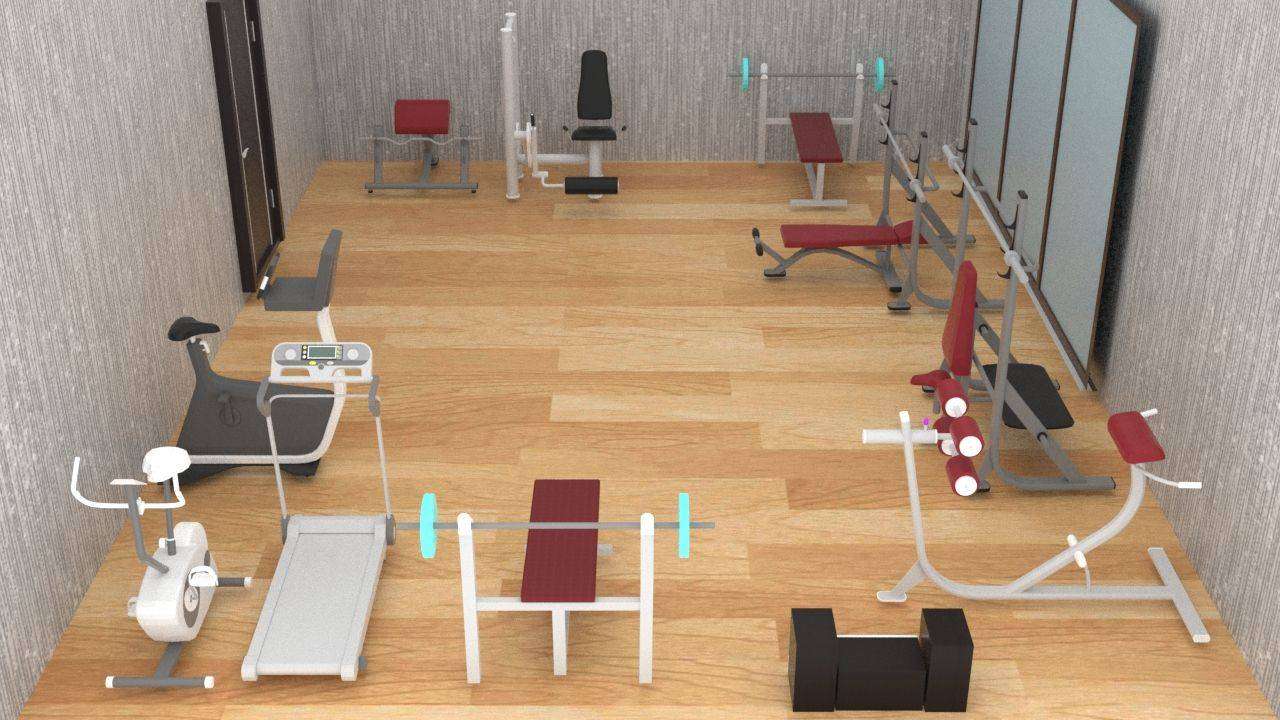} \\
            \includegraphics[width=0.195\linewidth]{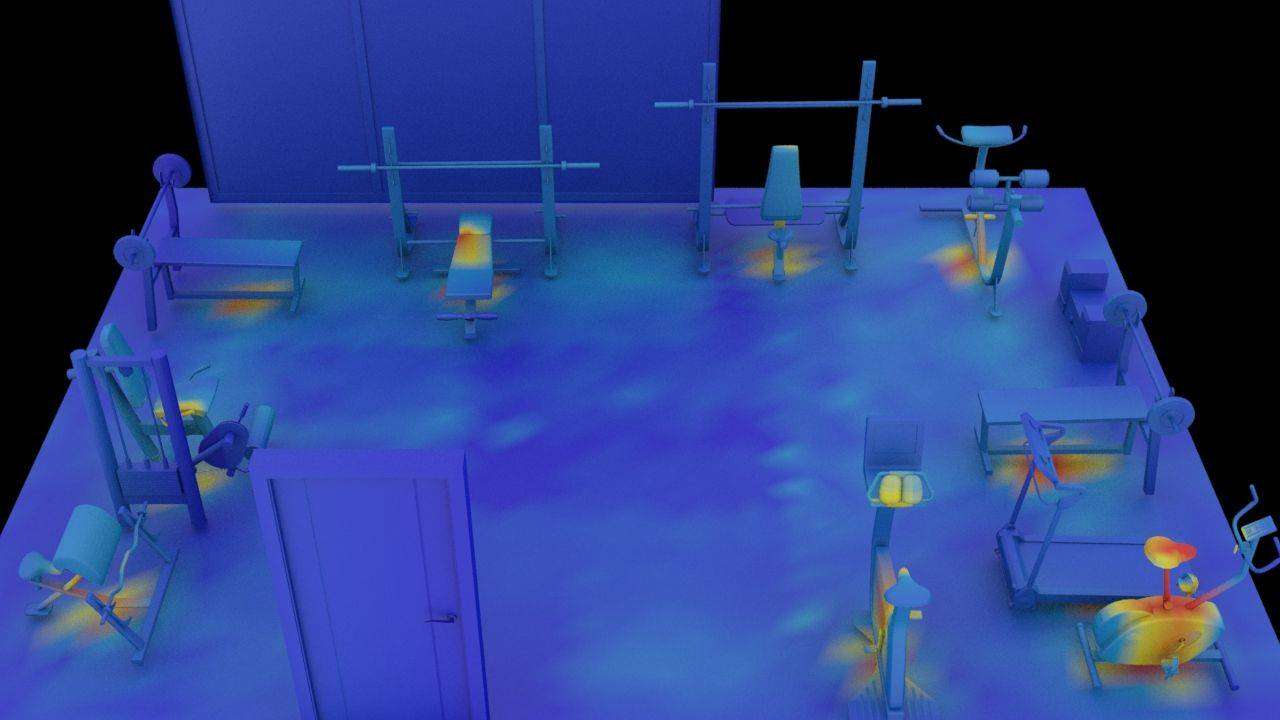}
        \end{tabular}}\hfill
        &
        \subfloat[kitchen]{
        \begin{tabular}[b]{c}
            \includegraphics[width=0.195\linewidth]{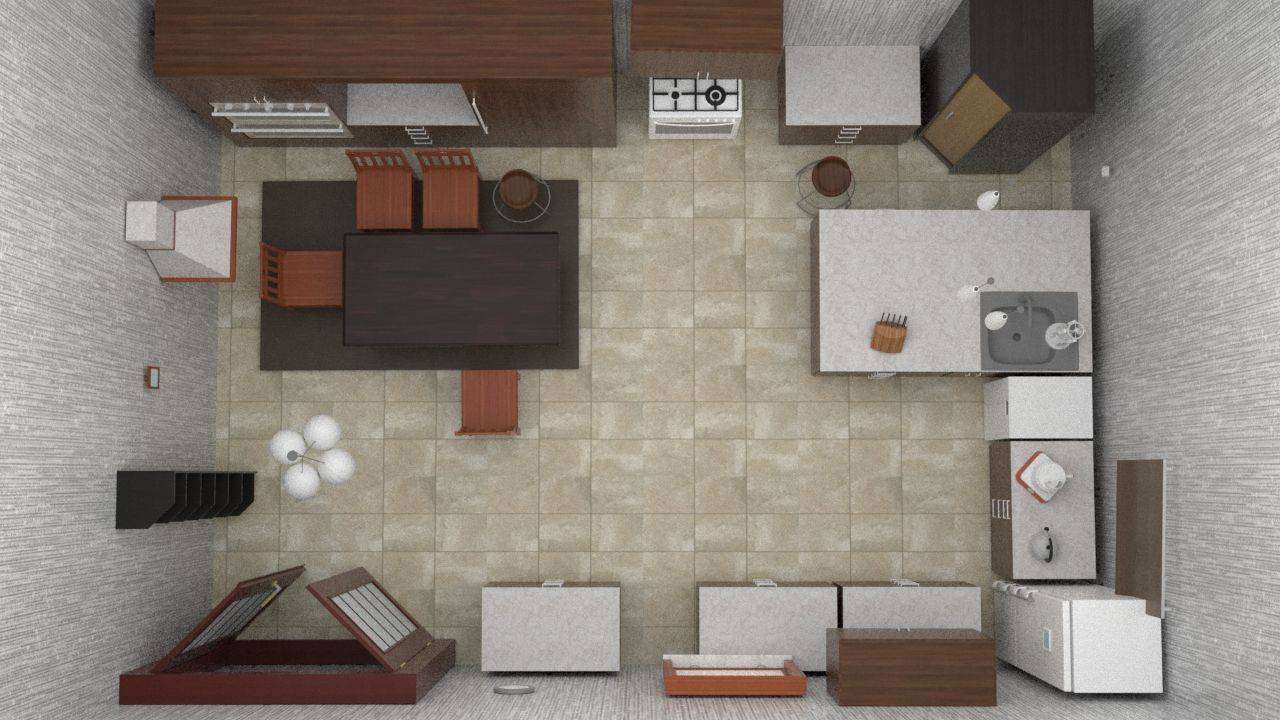} \\
            \includegraphics[width=0.195\linewidth]{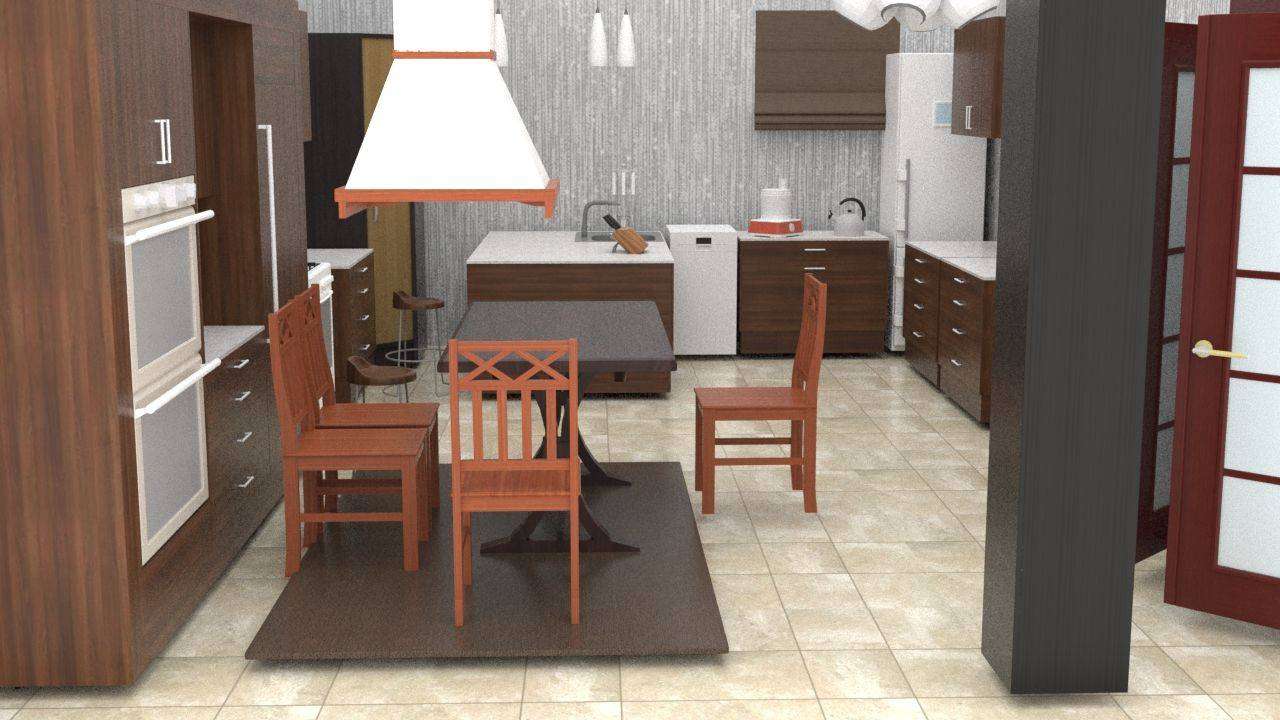} \\
            \includegraphics[width=0.195\linewidth]{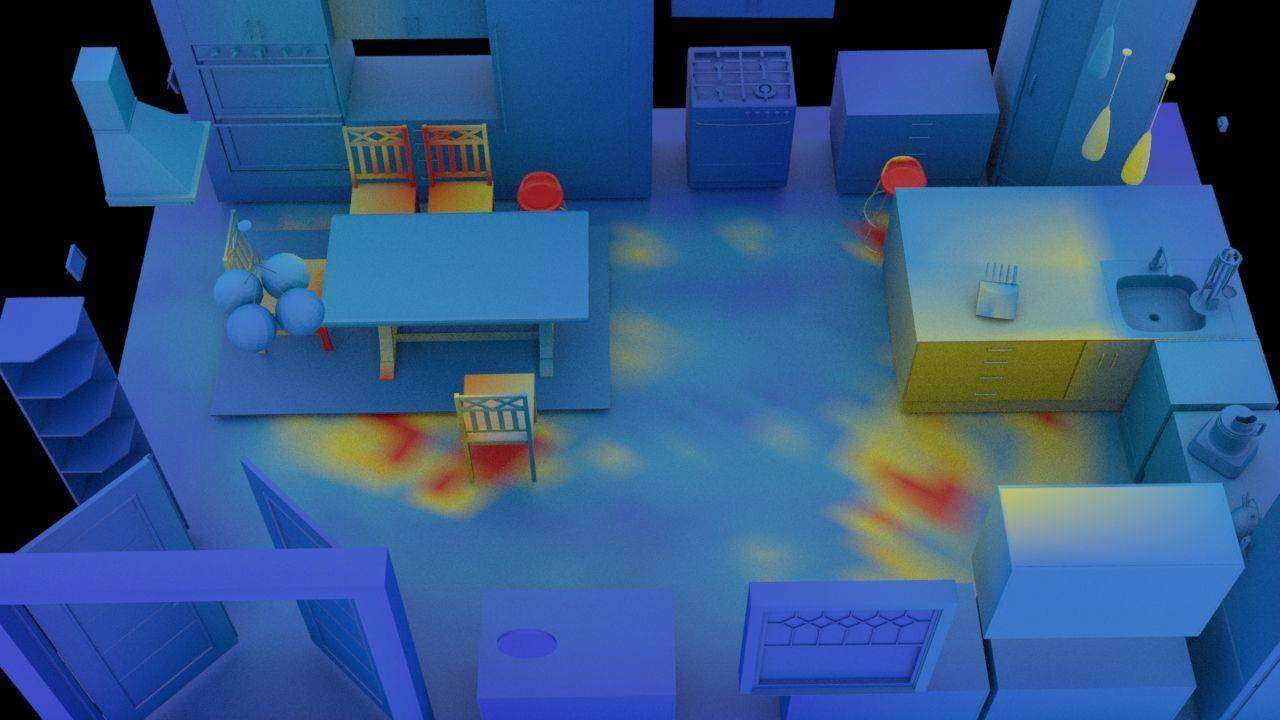}
        \end{tabular}}\hfill
        &
        \subfloat[living room]{
        \begin{tabular}[b]{c}
            \includegraphics[width=0.195\linewidth]{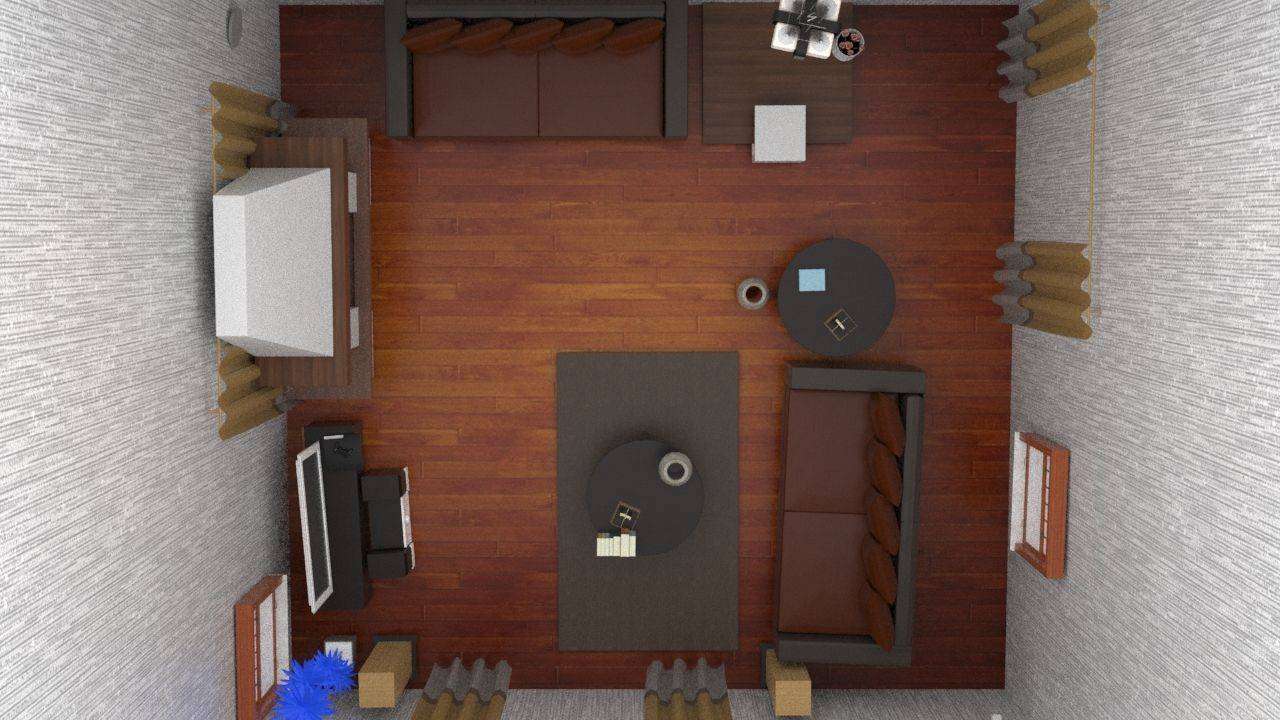} \\
            \includegraphics[width=0.195\linewidth]{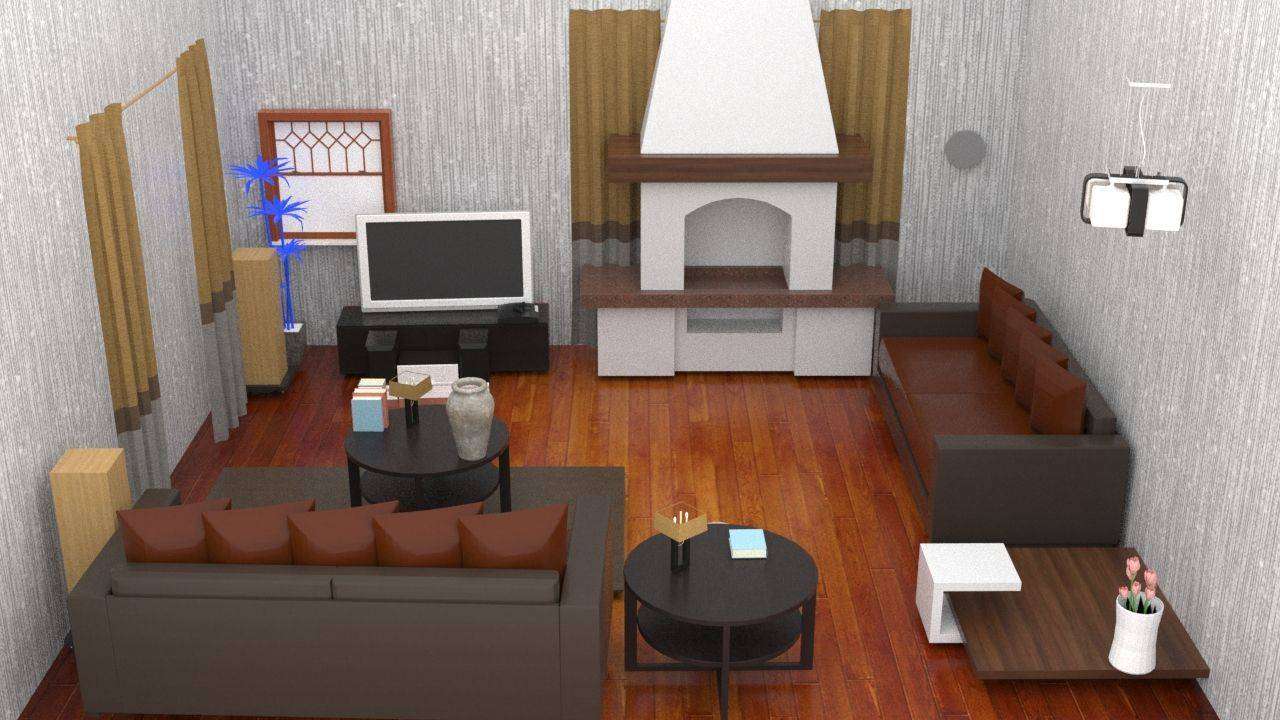} \\
            \includegraphics[width=0.195\linewidth]{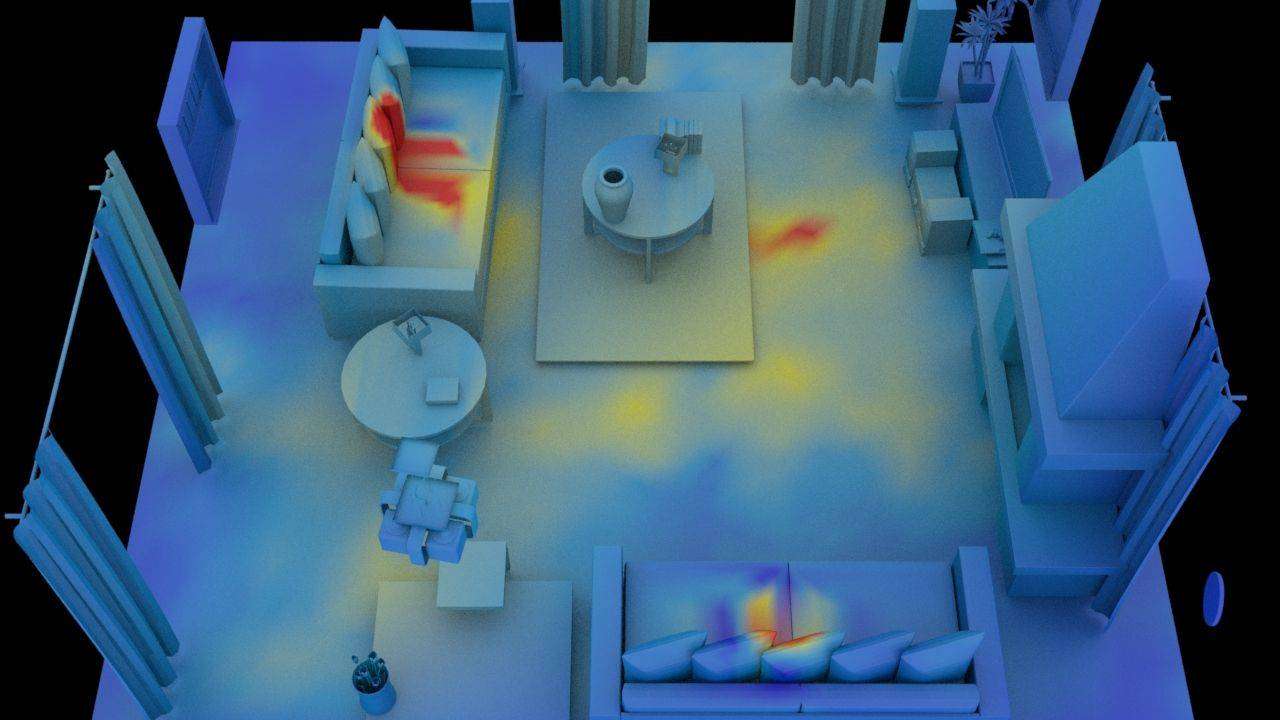}
        \end{tabular}}\hfill
        &
        \subfloat[office]{
        \begin{tabular}[b]{c}
            \includegraphics[width=0.195\linewidth]{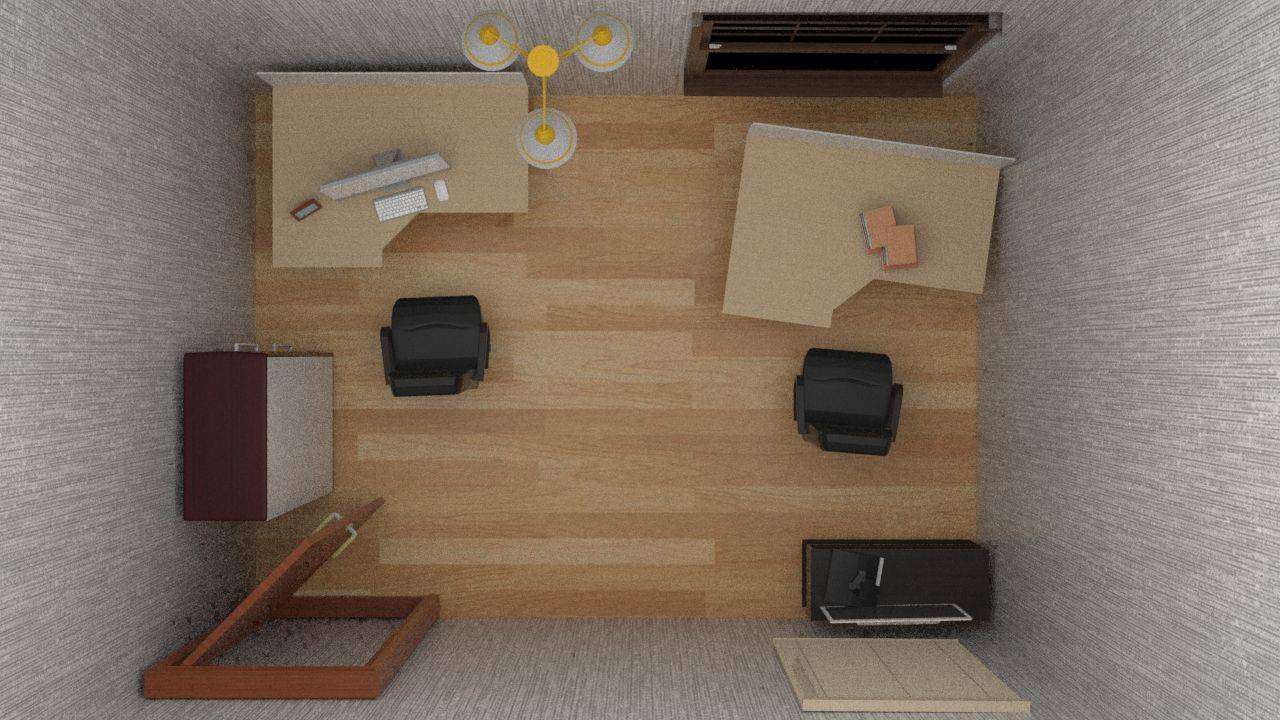} \\
            \includegraphics[width=0.195\linewidth]{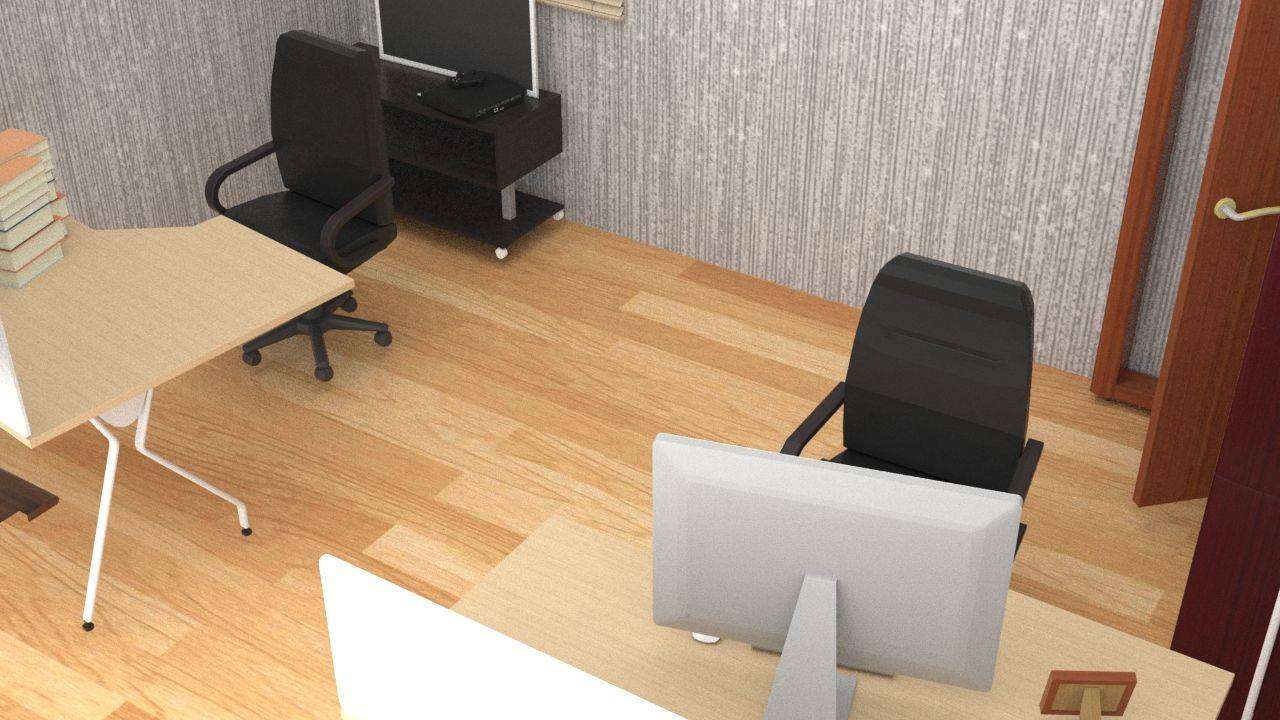} \\
            \includegraphics[width=0.195\linewidth]{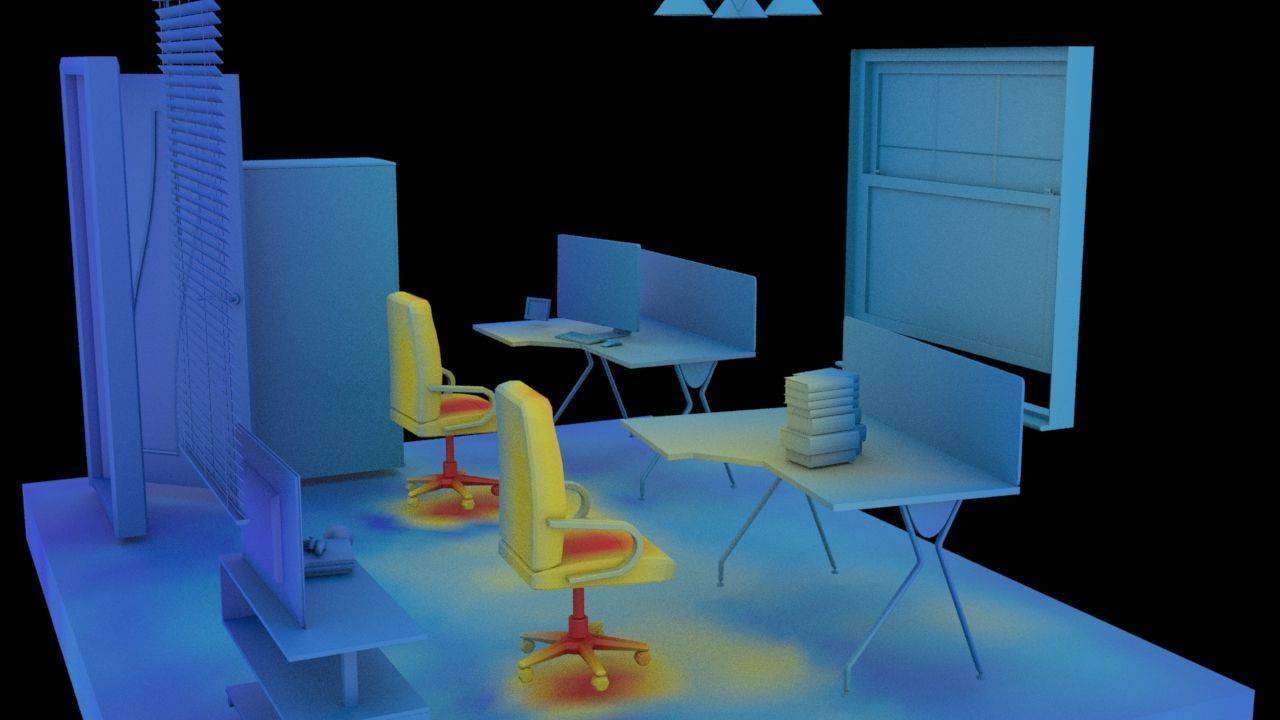}
        \end{tabular}}\hfill
        &
        \subfloat[storage]{
        \begin{tabular}[b]{c}
            \includegraphics[width=0.195\linewidth]{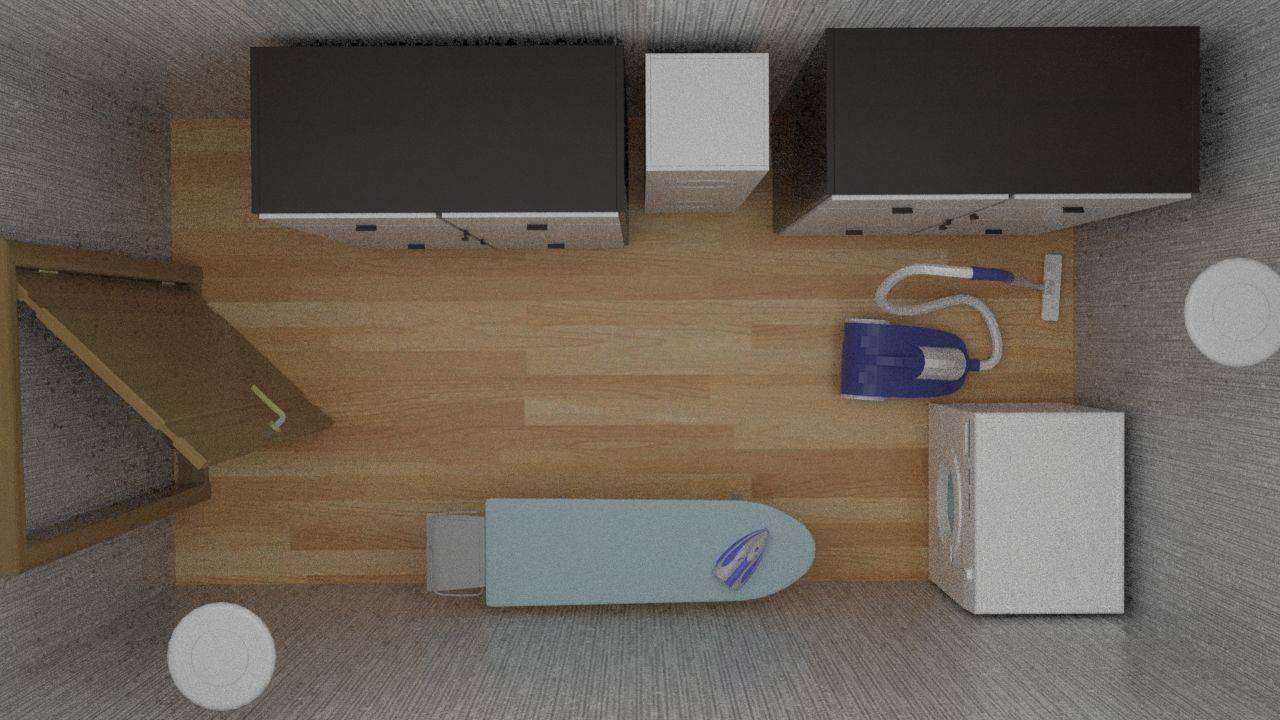} \\
            \includegraphics[width=0.195\linewidth]{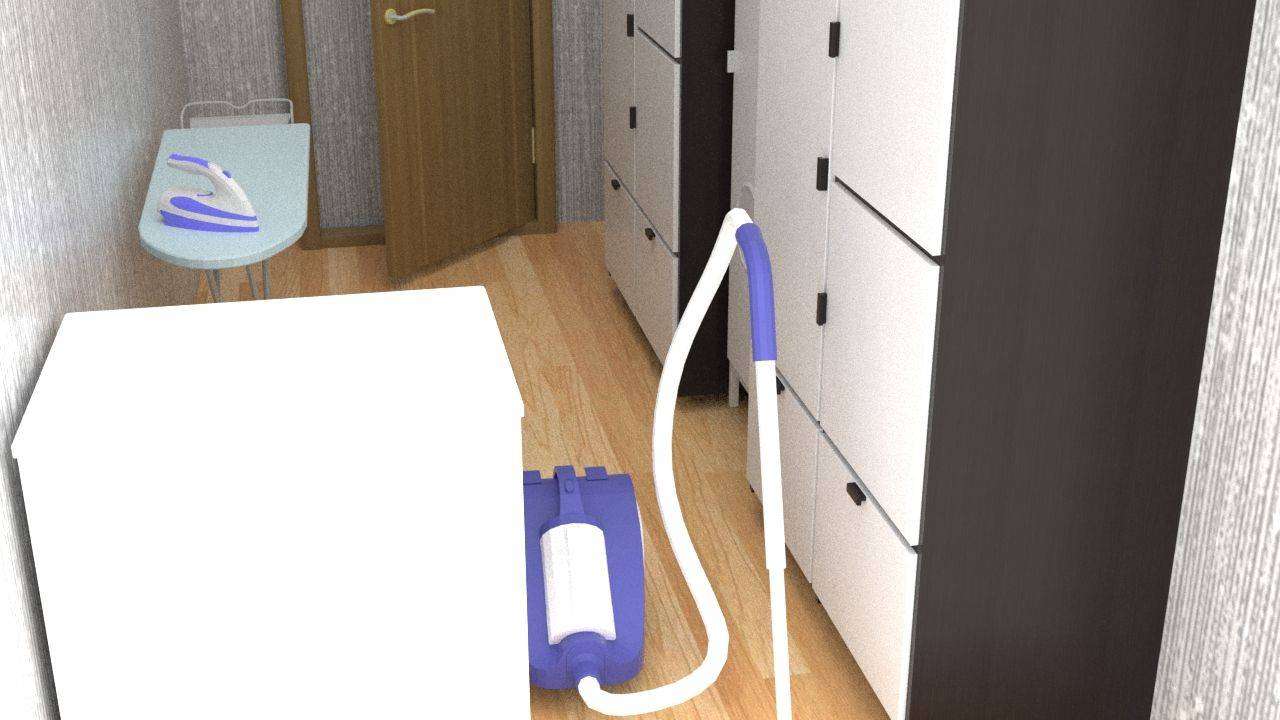} \\
            \includegraphics[width=0.195\linewidth]{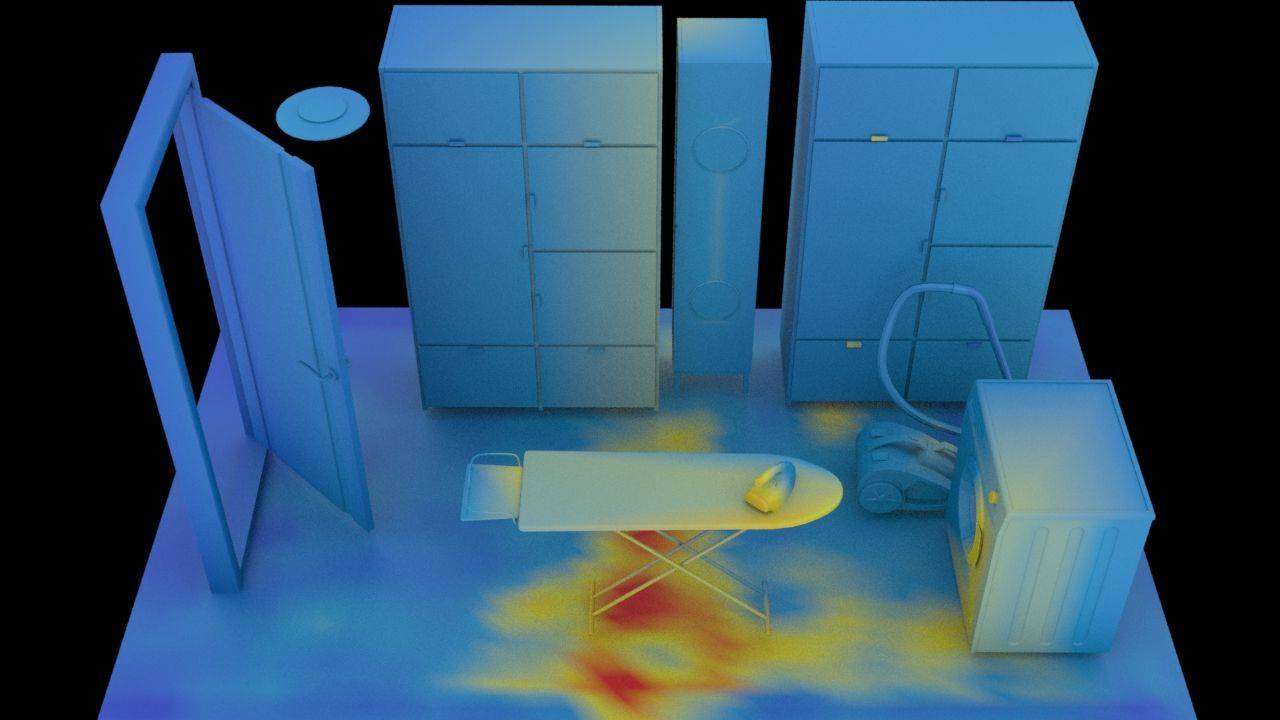}
        \end{tabular}}\hfill \\
    \end{tabular}
    \vspace{-5pt}
    \caption{Examples of scenes in ten different categories. Top: top-view. Middle: a side-view. Bottom: affordance heatmap.}
    \vspace{-18pt}
    \label{fig:results}
\end{center}
\end{figure*}

\begin{table}[b]
\begin{center}
\vspace{-15pt}
\caption{Classification results on segmentation maps of synthesized scenes using different methods vs. SUNCG.}
\vspace{-9pt}
\resizebox{0.48\textwidth}{!}{
\begin{tabular}{c|c|c|c}
 \Xhline{2\arrayrulewidth}
 \textbf{Method} & Yu \etal~\cite{yu2011make} & SUNCG Perturbed & Ours \\
 \hline
 \textbf{Accuracy(\%) $\downarrow$} & 87.49 & 63.69 & 76.18 \\
 \Xhline{2\arrayrulewidth}
\end{tabular}
}
\label{tab:classify_result}
\vspace{-13pt}
\end{center}
\end{table}

\begin{table*}[ht]
\begin{center}
\caption{Comparison between affordance maps computed from our samples and real data}
\vspace{-9pt}
\label{tab:affordance_comparison}
\bgroup
\def\arraystretch{1.0}
\setlength\tabcolsep{0.2em}
\resizebox{0.85\textwidth}{!}{
\begin{tabular}{l|c|c|c|c|c|c|c|c|c|c}

\Xhline{2\arrayrulewidth}
\bf{Metric}        & Bathroom      & Bedroom      & Dining Room   & Garage       & Guest Room   & Gym          & Kitchen      & Living Room   & Office       & Storage      \\
\hline
 Total variation & 0.431 & 0.202 & 0.387 & 0.237 & 0.175 & 0.278 & 0.227 & 0.117 & 0.303 & 0.708 \\
 Hellinger distance & 0.453 & 0.252 & 0.442 & 0.284 & 0.212 & 0.294 & 0.251 & 0.158  & 0.318 & 0.703 \\
\Xhline{2\arrayrulewidth}
\end{tabular}
}
\egroup
\vspace{-14pt}
\end{center}
\end{table*}

\begin{table*}[ht]
\begin{center}
\caption{Human subjects' ratings (1-5) of the sampled layouts based on functionality (top) and naturalness (bottom)}
\vspace{-9pt}
\label{tab:human_evaluation}
\bgroup
\def\arraystretch{1.0}
\setlength\tabcolsep{0.2em}
\resizebox{\textwidth}{!}{
\begin{tabular}{l|c|c|c|c|c|c|c|c|c|c}

\Xhline{2\arrayrulewidth}
\bf{Method}        & Bathroom     & Bedroom      & Dining Room   & Garage       & Guest Room   & Gym          & Kitchen      & Living Room   & Office       & Storage      \\
\hline
 no-context & 1.12 $\pm$ 0.33 & 1.25 $\pm$ 0.43 & 1.38 $\pm$ 0.48  & 1.75 $\pm$ 0.66 & 1.50 $\pm$ 0.50 & 3.75 $\pm$ 0.97 & 2.38 $\pm$ 0.48 & 1.50 $\pm$ 0.87  & 1.62 $\pm$ 0.48 & 1.75 $\pm$ 0.43 \\
 object & 3.12 $\pm$ 0.60 & 3.62 $\pm$ 1.22 & 2.50 $\pm$ 0.71  & 3.50 $\pm$ 0.71 & 2.25 $\pm$ 0.97 & 3.62 $\pm$ 0.70 & 3.62 $\pm$ 0.70 & 3.12 $\pm$ 0.78  & 1.62 $\pm$ 0.48 & 4.00 $\pm$ 0.71 \\
 Yu~\etal~\cite{yu2011make} & 3.61 $\pm$ 0.52 & 4.15 $\pm$ 0.25 & 3.15 $\pm$ 0.40  & 3.59 $\pm$ 0.51 & 2.58 $\pm$ 0.31 & 2.03 $\pm$ 0.56 & 3.91 $\pm$ 0.98 & 4.62 $\pm$ 0.21  & 3.32 $\pm$ 0.81 & 2.58 $\pm$ 0.64 \\
 ours   & 4.58 $\pm$ 0.86 & 4.67 $\pm$ 0.90 & 3.33 $\pm$ 0.90  & 3.96 $\pm$ 0.79 & 3.25 $\pm$ 1.36 & 4.04 $\pm$ 0.79 & 4.21 $\pm$ 0.87 & 4.58 $\pm$ 0.86  & 3.67 $\pm$ 0.75 & 4.79 $\pm$ 0.58 \\
\hline
 no-context & 1.00 $\pm$ 0.00 & 1.00 $\pm$ 0.00 & 1.12 $\pm$ 0.33  & 1.38 $\pm$ 0.70 & 1.12 $\pm$ 0.33 & 1.62 $\pm$ 0.86 & 1.00 $\pm$ 0.00 & 1.25 $\pm$ 0.43  & 1.12 $\pm$ 0.33 & 1.00 $\pm$ 0.00 \\
 object & 2.88 $\pm$ 0.78 & 3.12 $\pm$ 1.17 & 2.38 $\pm$ 0.86  & 3.00 $\pm$ 0.71 & 2.50 $\pm$ 0.50 & 3.38 $\pm$ 0.86 & 3.25 $\pm$ 0.66 & 2.50 $\pm$ 0.50  & 1.25 $\pm$ 0.43 & 3.75 $\pm$ 0.66 \\
 Yu~\etal~\cite{yu2011make} & 4.00 $\pm$ 0.52 & 3.85 $\pm$ 0.92 & 3.27 $\pm$ 1.01  & 2.99 $\pm$ 0.25 & 3.52 $\pm$ 0.93 & 2.14 $\pm$ 0.63 & 3.89 $\pm$ 0.90 & 3.31 $\pm$ 0.29  & 2.77 $\pm$ 0.67 & 2.96 $\pm$ 0.41 \\
 ours   & 4.21 $\pm$ 0.71 & 4.25 $\pm$ 0.66 & 3.08 $\pm$ 0.70  & 3.71 $\pm$ 0.68 & 3.83 $\pm$ 0.80 & 4.17 $\pm$ 0.75 & 4.38 $\pm$ 0.56 & 3.42 $\pm$ 0.70  & 3.25 $\pm$ 0.72 & 4.54 $\pm$ 0.71 \\
\Xhline{2\arrayrulewidth}
\end{tabular}
}
\egroup
\vspace{-20pt}
\end{center}
\end{table*}

\section{Experiments}\label{sec:experiments}
We design three experiments based on different criteria: i) visual similarity to manually constructed scenes, ii) the accuracy of affordance maps for the synthesized scenes, and iii) functionalities and naturalness of the synthesized scenes. The first experiment compares our method with a state-of-the-art room arrangement method; the second experiment measures the synthesized affordances; the third one is an ablation study. Overall, the experiments show that our algorithm can robustly sample a large variety of realistic scenes that exhibits naturalness and functionality.

\begin{figure}[t!]
\begin{center}
    \vspace{-8pt}
    \begin{tabular}[c]{@{\hskip-1.4em}c@{\hskip-1em}c@{\hskip-1em}c@{\hskip-1em}}
        \captionsetup[subfigure]{aboveskip=-6pt,belowskip=-6pt}
        \subfloat[SUNCG Perturbed]{
        \begin{tabular}[b]{c}
            \includegraphics[width=0.32\linewidth]{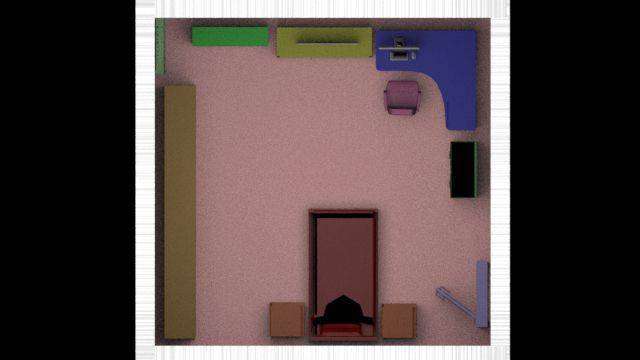}
        \end{tabular}}\hfill
        &
        \subfloat[Yu \etal~\cite{yu2011make}]{
        \begin{tabular}[b]{c}
            \includegraphics[width=0.32\linewidth]{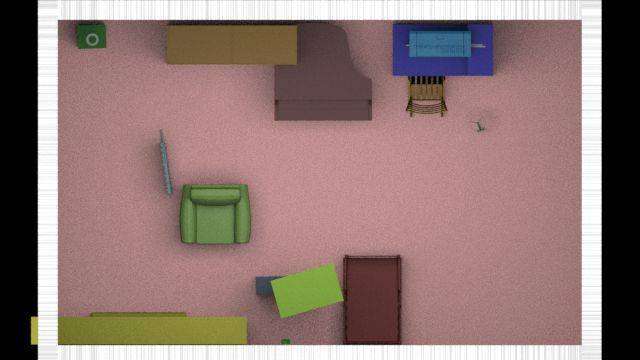}
        \end{tabular}}\hfill
        &
        \subfloat[Ours]{
        \begin{tabular}[b]{c}
            \includegraphics[width=0.32\linewidth]{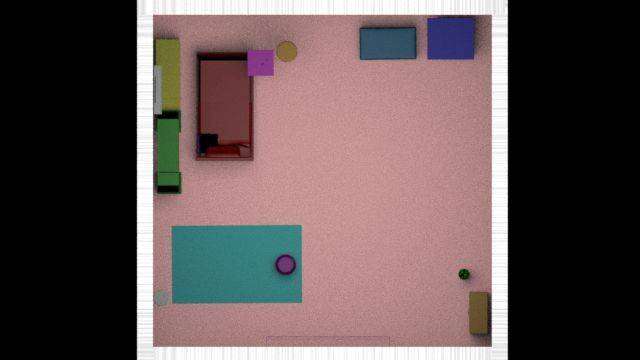}
        \end{tabular}}\hfill
    \end{tabular}
    \vspace{-8pt}
    \caption{Top-view segmentation maps for classification.}
    \vspace{-25pt}
    \label{fig:segmentation}
\end{center}
\end{figure}

\begin{figure}[t!]
\begin{center}
	\vspace{-8pt}
    \begin{tabular}[c]{@{\hskip-1.4em}c@{\hskip-1em}c@{\hskip-1em}c@{\hskip-1em}}
        \captionsetup[subfigure]{aboveskip=-6pt,belowskip=-6pt}
        \subfloat{
        \begin{tabular}[b]{c}
            \includegraphics[width=0.32\linewidth]{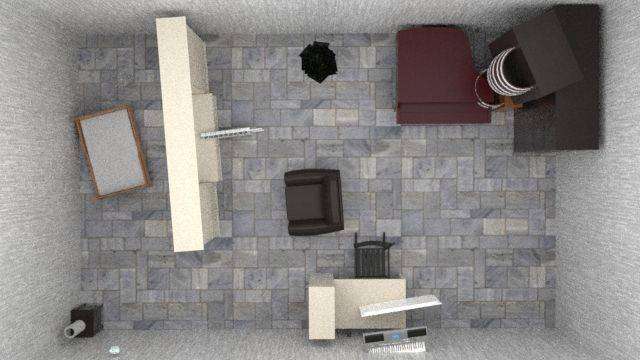} \\
            \includegraphics[width=0.32\linewidth]{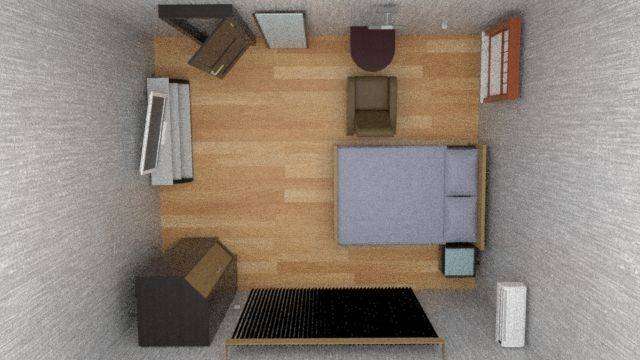}
        \end{tabular}}\hfill
        &
        \subfloat{
        \begin{tabular}[b]{c}
            \includegraphics[width=0.32\linewidth]{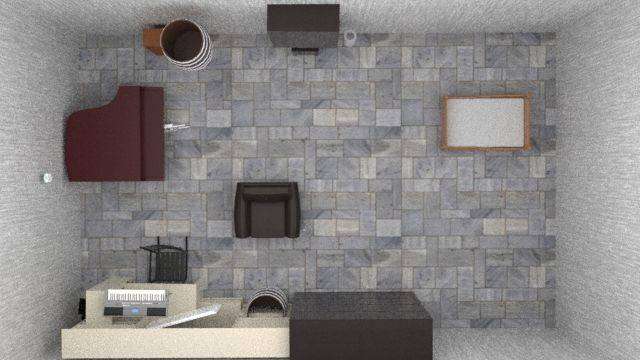} \\
            \includegraphics[width=0.32\linewidth]{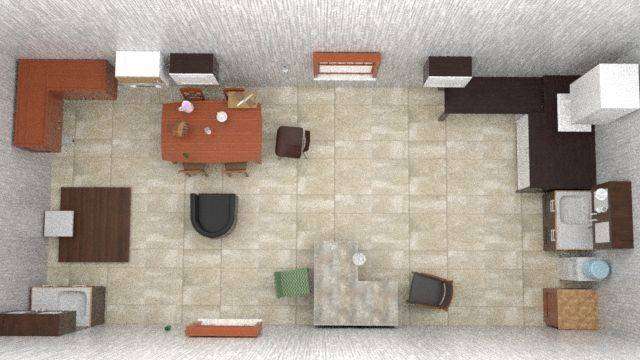}
        \end{tabular}}\hfill
        &
        \subfloat{
        \begin{tabular}[b]{c}
            \includegraphics[width=0.32\linewidth]{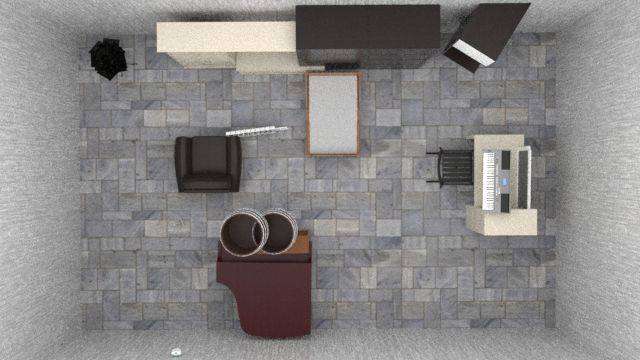} \\
            \includegraphics[width=0.32\linewidth]{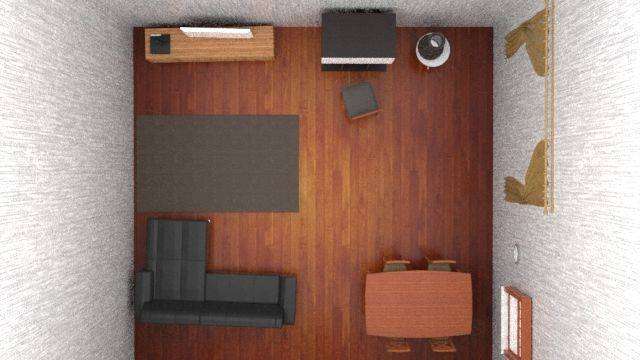}
        \end{tabular}}\hfill
    \end{tabular}
	\vspace{-5pt}
    \caption{\textbf{Top}: previous methods~\cite{yu2011make} only re-arranges a given input scene with a fixed room size and a predefined set of objects. \textbf{Bottom}: our method samples a large variety of scenes.}
	\vspace{-20pt}
    \label{fig:comparison}
\end{center}
\end{figure}

\noindent\textbf{Layout Classification.} To quantitatively evaluate the visual realism, we trained a classifier on the top-view segmentation maps of synthesized scenes and SUNCG scenes. Specifically, we train a ResNet-152~\cite{he2016deep} to classify top view layout segmentation maps (synthesized vs. SUNCG). Examples of top-view segmentation maps are shown in Figure~\ref{fig:segmentation}. The reason to use segmentation maps is that we want to evaluate the room layout excluding rendering factors such as object materials. We use two methods for comparison: i) a state-of-the-art furniture arrangement optimization method proposed by Yu \etal~\cite{yu2011make}, and ii) slight perturbation of SUNCG scenes by adding small Gaussian noise (\eg $\mu = 0, \sigma = 0.1$) to the layout. The room arrangement algorithm proposed by~\cite{yu2011make} takes one pre-fixed input room and re-organizes the room. 1500 scenes are randomly selected for each method and SUNCG: 800 for training, 200 for validation, and 500 for testing. As shown in Table~\ref{tab:classify_result}, the classifier successfully distinguishes Yu \etal vs. SUNCG with an accuracy of 87.49\%. Our method achieves a better performance of 76.18\%, exhibiting a higher realism and larger variety. This result indicates our method is much more visually similar to real scenes than the comparative scene optimization method. Qualitative comparisons of Yu \etal and our method are shown in Figure~\ref{fig:comparison}.

\noindent\textbf{Affordance Maps Comparison.} We sample 500 rooms of 10 different scene categories summarized in Table~\ref{tab:affordance_comparison}. For each type of room, we compute the affordance maps of the objects in the synthesized samples, and calculate both the total variation distances and Hellinger distances between the affordance maps computed from the synthesized samples and the SUNCG dataset. The two distributions are similar if the distance is close to 0. Most sampled scenes using the proposed method show similar affordance distributions to manually created ones from SUNCG. Some scene types (\eg Storage) show a larger distance since they do not exhibit clear affordances. Overall, the results indicate that affordance maps computed from the synthesized scenes are reasonably close to the ones computed from manually constructed scenes by artists.

\noindent\textbf{Functionality and naturalness.} Three methods are used for comparison: (i) direct sampling of rooms according to the statistics of furniture occurrence without adding contextual relation, (ii) an approach that only models object-wise relations by removing the human constraints in our model, and (iii) the algorithm proposed by Yu \etal~\cite{yu2011make}. We showed the sampled layouts using three methods to 4 human subjects. Subjects were told the room category in advance, and instructed to rate given scene layouts without knowing the method used to generate the layouts. For each of the 10 room categories, 24 samples were randomly selected using our method and ~\cite{yu2011make}, and 8 samples were selected using both the object-wise modeling method and the random generation. The subjects evaluated the layouts based on two criteria: (i) functionality of the rooms, \eg, can the ``bedroom" satisfies a human's needs for daily life; and (ii) the naturalness and realism of the layout. Scales of responses range from 1 to 5, with 5 indicating perfect functionalilty or perfect naturalness and realism. The mean ratings and the standard deviations are summarized in Table~\ref{tab:human_evaluation}. Our approach outperforms the three methods in both criteria, demonstrating the ability to sample a functionally reasonable and realistic scene layout. More qualitative results are shown in Figure~\ref{fig:results}.

\noindent\textbf{Complexity of synthesis.} The time complexity is hard to measure since MCMC sampling is adopted. Empirically, it takes about 20-40 minutes to sample an interior layout (20000 iterations of MCMC), and roughly 12-20 minutes to render a 640$\times$480 image on a normal PC. The rendering speed depends on settings related to illumination, environments, and the size of the scene, \etc.

\section{Conclusion}
We propose a novel general framework for human-centric indoor scene synthesis by sampling from a spatial And-Or graph. The experimental results demonstrate the effectiveness of our approach over a large variety of scenes based on different criteria. In the future, to synthesize physically plausible scenes, a physics engine should be integrated. We hope the synthesized data can contribute to the broad AI community.

%

\newpage

\section*{Acknowledgment}

The authors thank Professor Ying Nian Wu from UCLA Statistics Department and Professor Demetri Terzopoulos from UCLA Computer Science Department for insightful discussions.
The work reported herein is supported by DARPA XAI N66001-17-2-4029 and ONR MURI N00014-16-1-2007.

{\small
\bibliographystyle{ieee}
\bibliography{egbib}
}

\clearpage
\title{Supplementary Material for\\
Human-centric Indoor Scene Synthesis Using Stochastic Grammar}

\author{
Siyuan Qi$^{1}$ \quad Yixin Zhu$^{1}$ \quad Siyuan Huang$^{1}$ \quad Chenfanfu Jiang$^{2}$ \quad  Song-Chun Zhu$^1$\\[12pt]1
$^1$ UCLA Center for Vision, Cognition, Learning and Autonomy\\
$^2$ UPenn Computer Graphics Group
}

\maketitle

\section{Simulated Annealing}
The simulated annealing schedule is important for synthesizing realistic scenes. In our experiments, we set the total sampling iterations to 20000, and it takes around 20 minutes to sample an interior layout. We use the following simulated schedule for sampling:
\begin{equation}
\begin{aligned}
T(t) = \frac{T_0}{\ln (1+t)}
\end{aligned}
\end{equation}
where $T(t)$ is the temperature at iteration $t$. Geman \etal~\cite{geman1984stochastic} proved that $T(t) \geq \frac{T_0}{\ln (1+t)}$ is a necessary and sufficient condition to ensure convergence to the global minimum with probability one.

\begin{table*}[t]
    \caption{Depth estimation with different training protocols.}
   \centering
   \resizebox{0.9\textwidth}{!}{
   \renewcommand{\arraystretch}{1.2}
	\setlength\tabcolsep{6pt}
	\begin{tabular}{|c c|c c c c c|c c c|}
	\hline
	\multirow{3}{*}{pre-Train} & \multirow{3}{*}{fine-Tune} &
	\multicolumn{5}{c|}{Error} &
	\multicolumn{3}{c|}{Accuracy} \\ 
	\cline{3-10}
	& & Abs Rel & Sqr Rel & Log10 & RMSE(linear) & RMSE(log) & $\delta < 1.25 $ & $\delta < 1.25^2$ & $\delta < 1.25^3$\\
	\hline
	NYUv2 & - & 0.233 & 0.158 & 0.098 & 0.831 & 0.117 & 0.605 & 0.879 & 0.965\\
	Ours & - & 0.241 & 0.173 & 0.108 & 0.842 & 0.125 & 0.612 & 0.882 & 0.966\\
	Ours & NYUv2 & \textbf{0.226} & \textbf{0.152} & \textbf{0.090} & \textbf{0.820} & \textbf{0.108} & \textbf{0.616} & \textbf{0.887} & \textbf{0.972}\\ \hline
	\end{tabular}
	}
    \label{table:depth_comparison}
\end{table*}

\begin{table}
\caption{Normal estimation with different training protocols.}
\centering
		\resizebox{0.45\textwidth}{!}{
		\renewcommand{\arraystretch}{1.2}
		\setlength\tabcolsep{2pt}
 		\begin{tabular}{|c c|c c c c c|}
 		\hline
		pre-train & fine-tune & mean$\downarrow$ & median$\downarrow$ & $11.25^\circ\uparrow $ & $22.5^\circ\uparrow$ & $30^\circ\uparrow$\\
		\hline
		\multicolumn{2}{|c|}{NYUv2} & 27.30 & 21.12 & 27.21 & 52.61 & 64.72 \\
		\hline
		\multicolumn{2}{|c|}{Eigen~\cite{eigen2015predicting}} & 22.2 & 15.3 & 38.6 & 64.0 & 73.9 \\
		\hline
		~\cite{zhang2016physically} & NYUv2 & 21.74 & 14.75 & 39.37 & 66.25 & 76.06 \\
		\hline
		ours+\cite{zhang2016physically} & NYUv2 & \textbf{21.47} & \textbf{14.45} & \textbf{39.84} & \textbf{67.05} & \textbf{76.72} \\
		\hline
		\end{tabular}
		}
		\label{table:trained_normal}
\end{table}

\section{Data Effectiveness}
We further demonstrate that our data can be utilized to improve performance on two scene understanding tasks: depth estimation and surface normal estimation from single RGB images. We show that the performance of state-of-art methods can be improved when trained with our synthesized data along with natural images.

\paragraph{Depth estimation} 
Single-image depth estimation is a fundamental problem in computer vision, which has found broad applications in scene understanding, 3D modeling, and robotics. The problem is challenging since no reliable depth cues are available. In this task, the algorithms output a depth image based on a single RGB input image.

To demonstrate the efficacy of our synthetic data, we compare the depth estimation results provided by models trained following protocols similar to those we used in normal prediction with the network in~\cite{liu2015deep}. To perform a quantitative evaluation, we used the metrics applied in previous work~\cite{eigen2014depth}:
\begin{itemize}[leftmargin=*,noitemsep,nolistsep]
 \item Abs relative error:
 $\frac{1}{N}\sum_p\frac{\left|d_p-d_p^{gt}\right|}{d_p^{gt}}$,
 \item Square relative difference:
 $\frac{1}{N}\sum_p\frac{{\left|d_p-d_p^{gt}\right|}^2}{d_p^{gt}}$,
 \item Average $\log_{10}$ error: $\frac{1}{N}\sum_x{\left|
 \log_{10}(d_p)-\log_{10}(d_p^{gt})\right|}$,
 \item
 RMSE :$\sqrt{{\frac{1}{N}\sum_x{\left|d_p-d_p^{gt}\right|}}^2}$,
 \item
 Log RMSE:$\sqrt{{\frac{1}{N}\sum_x{\left|\log(d_p)-\log(d_p^{gt})\right|}}^2}$,
 \item Threshold: \% of $d_p \mbox{~s.t.}
 \max{(\frac{d_p}{d_p^{gt}},\frac{d_p^{gt}}{d_p}}) <
 \mbox{threshold}$,
\end{itemize}
where $d_p$ and $d_p^{gt}$ are the predicted depths and the ground truth depths at the pixel indexed by $p$, respectively, and $N$ is the number of pixels in all the evaluated images. The first five metrics capture the error calculated over all the pixels; lower values are better. The threshold criteria capture the estimation accuracy; higher values are better.

Table~\ref{table:depth_comparison} summarizes the results. We can see that the model pretrained on our dataset and fine-tuned on the NYU-Depth V2 dataset achieves the best performance, both in error and accuracy. This demonstrates the usefulness of our dataset in improving algorithm performance in scene understanding tasks.

\paragraph{Surface normal estimation}
Predicting surface normals from a single RGB image is an essential task in scene understanding since it provides important information in recovering the 3D structure of the scenes. We train a neural network with our synthetic data to demonstrate that the perfect per-pixel ground truth generated using our pipeline could be utilized to improve upon the state-of-the-art performance on a specific scene understanding task. Using the fully convolutional network model described by Zhang \etal~\cite{zhang2016physically}, we compare the normal estimation results given by models trained under two different protocols: (i) the network is directly trained and tested on the NYU-Depth V2 dataset, and (ii) the network is first pre-trained using our synthetic data, then fine-tuned and tested on NYU-Depth V2.

Following the standard evaluation protocol~\cite{fouhey2013data, bansal2016marr}, we evaluate a per-pixel error over the entire dataset. To evaluate the prediction error, we computed the mean, median, and RMSE of angular error between the predicted normals and ground truth normals. Prediction accuracy is given by calculating the fraction of pixels that are correct within a threshold $t$, where $t = 11.25^{\circ}, 22.5^{\circ},30^{\circ}$. Our experimental results are summarized in Table~\ref{table:trained_normal}. By utilizing our synthetic data, the model achieves better performance. The error mainly accrues in the area where the ground truth normal map is noisy. We argue that part of the reason is due to the sensor's noise or sensing distance limit. Such results in turn imply the importance to have perfect per-pixel ground truth for training and evaluation.

\section{More Qualitative Results}
See page 13-17.


\balance

\clearpage

\begin{figure*}[b!]
\begin{center}
\vspace{-36pt}
\includegraphics[width=\linewidth]{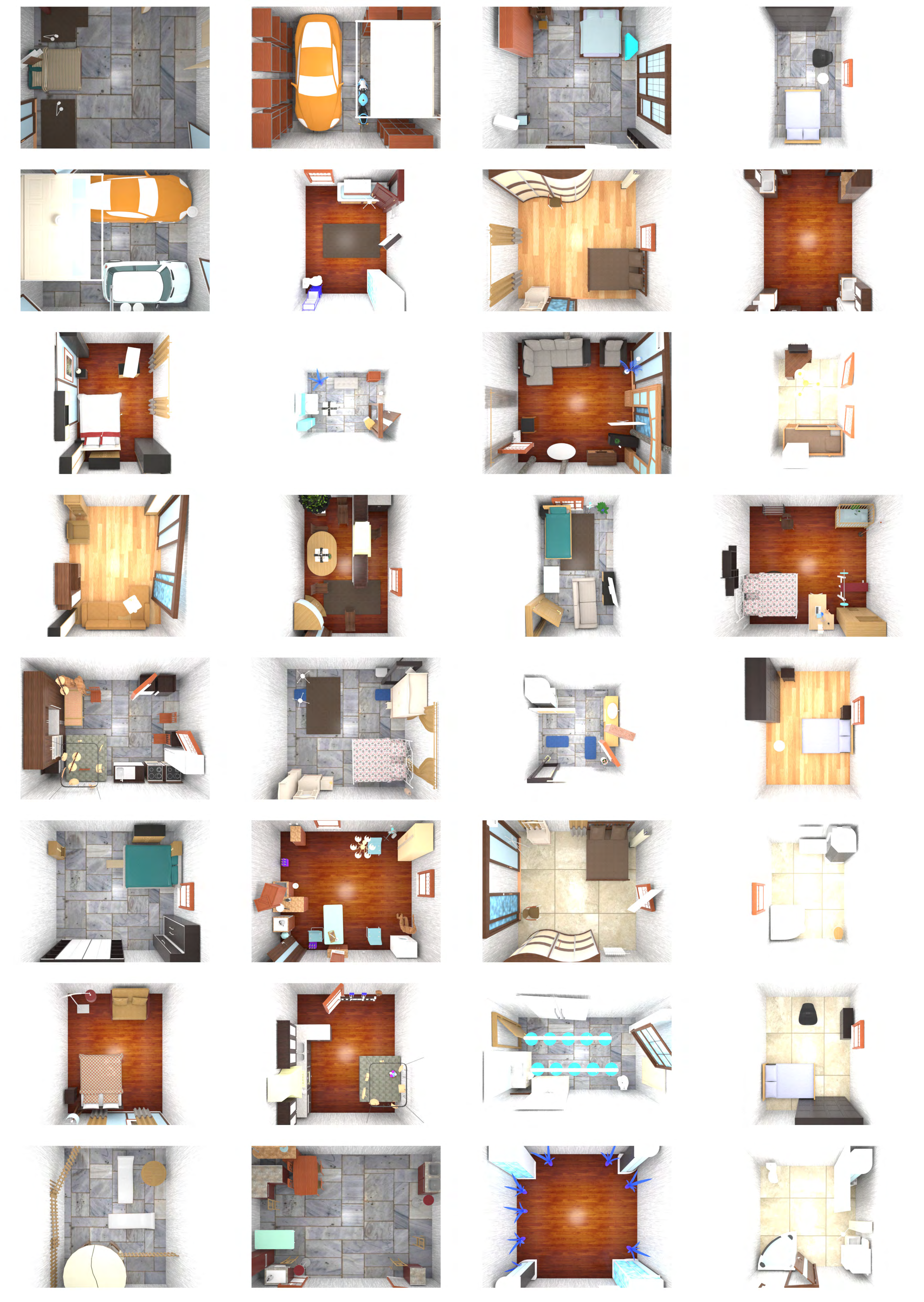}
\end{center}
\vspace{-15pt}
\end{figure*}

\clearpage

\begin{figure*}[b!]
\begin{center}
\vspace{-36pt}
\includegraphics[width=\linewidth]{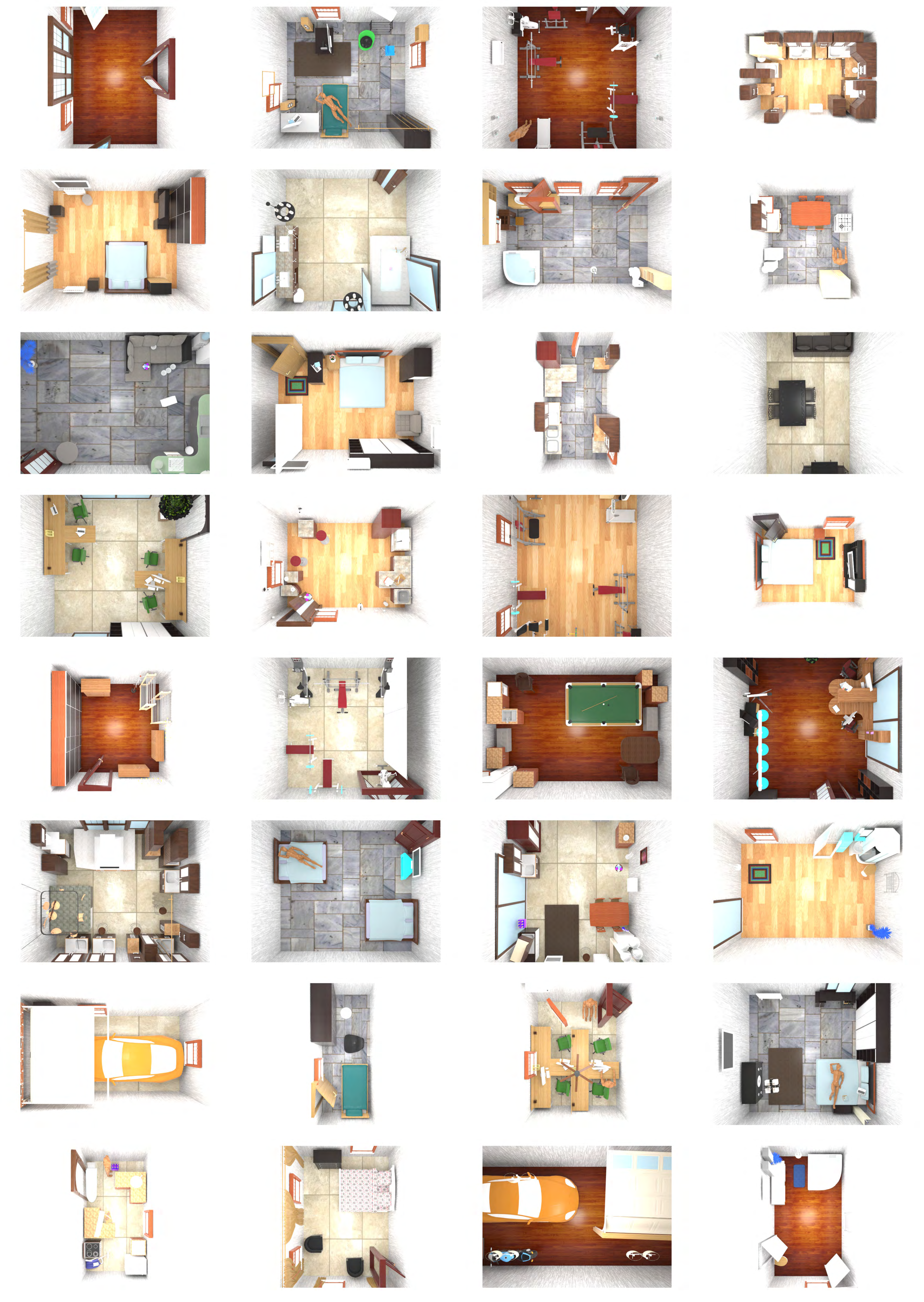}
\end{center}
\vspace{-15pt}
\end{figure*}

\clearpage

\begin{figure*}[b!]
\begin{center}
\vspace{-36pt}
\includegraphics[width=\linewidth]{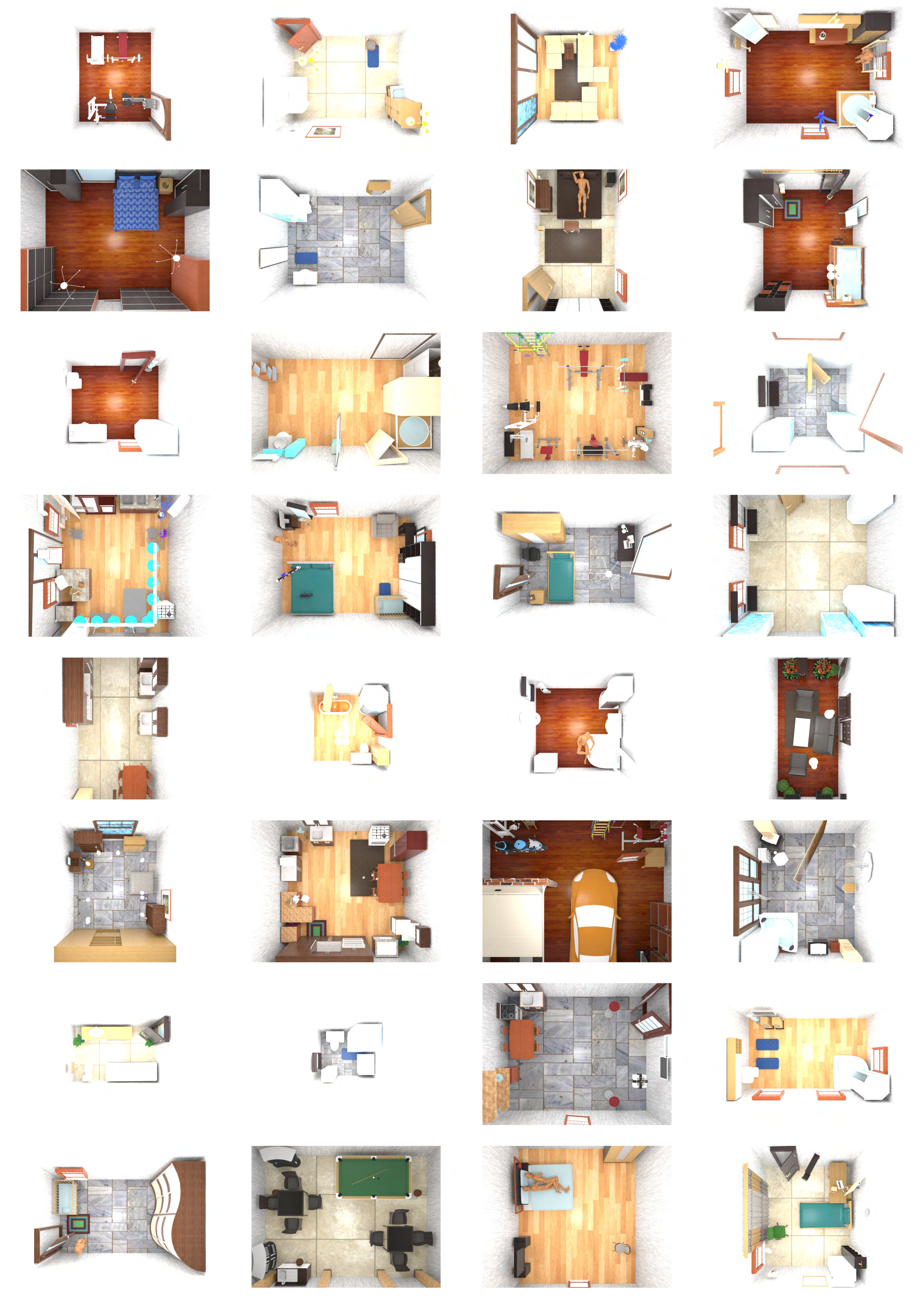}
\end{center}
\vspace{-15pt}
\end{figure*}

\clearpage

\begin{figure*}[b!]
\begin{center}
\vspace{-36pt}
\includegraphics[width=\linewidth]{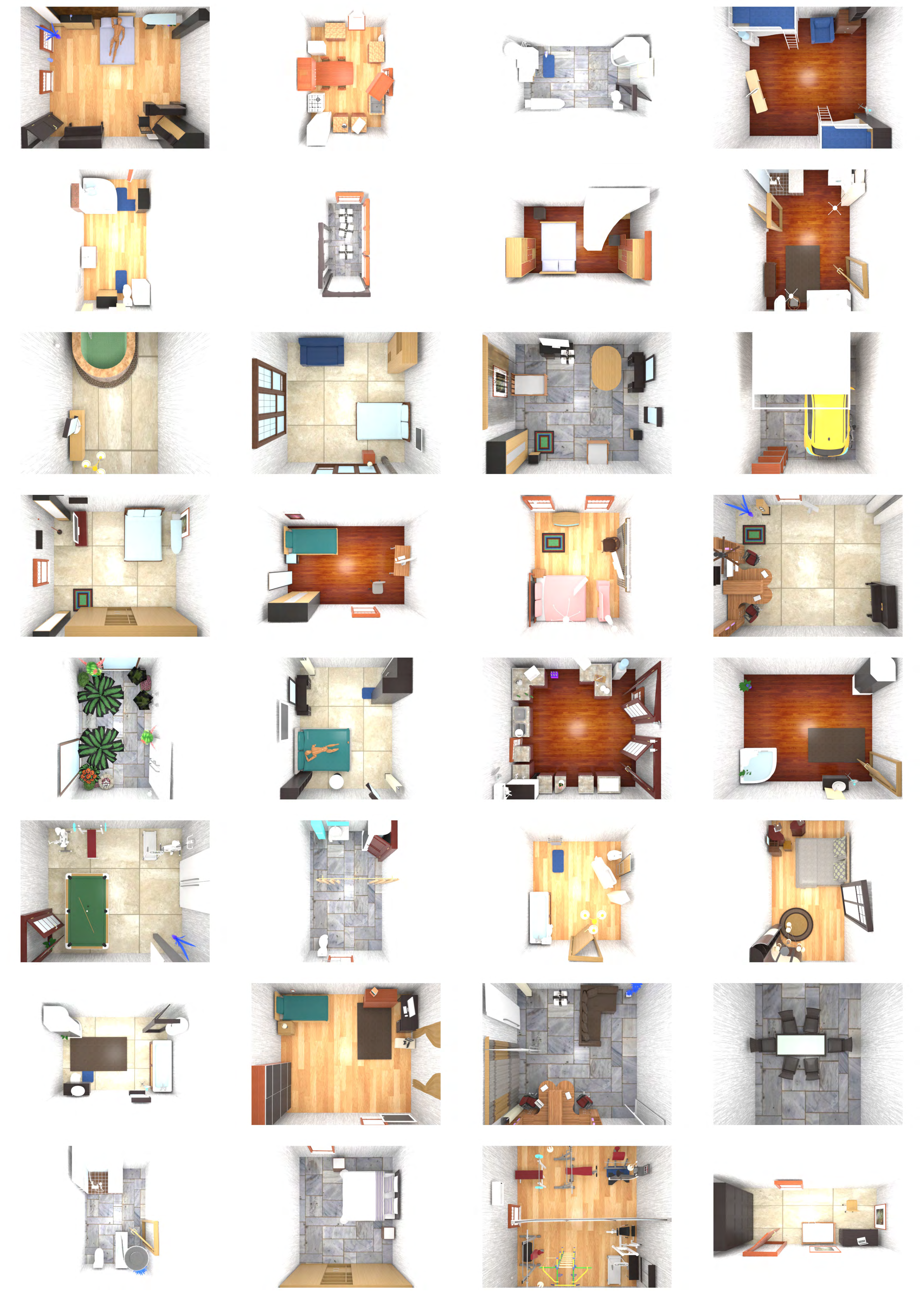}
\end{center}
\vspace{-15pt}
\end{figure*}

\begin{figure*}[b!]
\begin{center}
\vspace{-36pt}
\includegraphics[width=\linewidth]{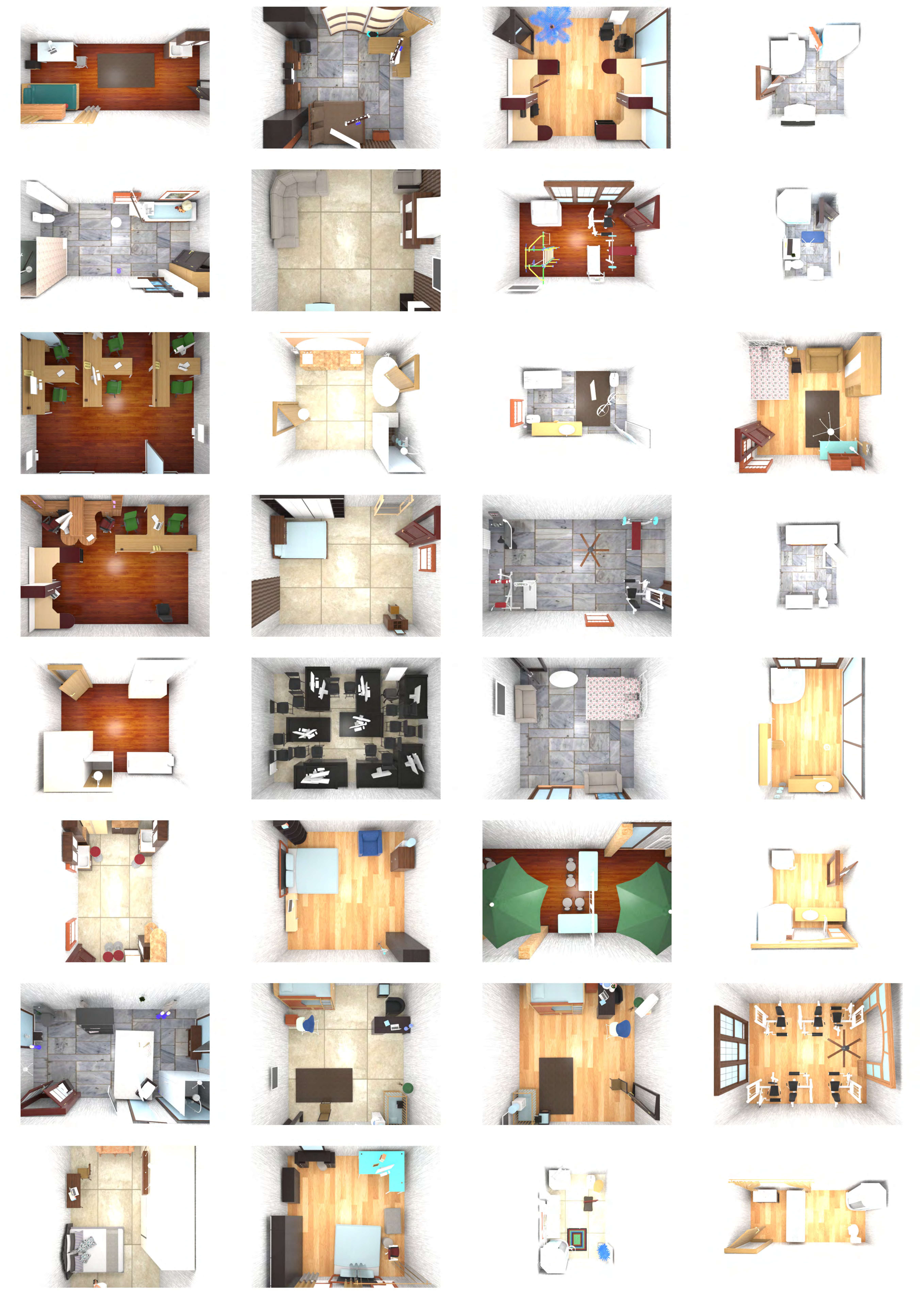}
\end{center}
\vspace{-15pt}
\end{figure*}

%

\end{document}